\documentclass[10pt,journal,compsoc]{IEEEtran}

\ifCLASSOPTIONcompsoc
  \usepackage[nocompress]{cite}
\else
  \usepackage{cite}
\fi

\ifCLASSINFOpdf

\else

\fi
\usepackage{algorithm}
\usepackage{algorithmic}
\usepackage{amsmath}
\usepackage{color}
\definecolor{gray}{rgb}{0.4,0.4,0.4}
\usepackage{epsfig}
\usepackage{subcaption}
\usepackage{enumitem}
\usepackage{array}
\usepackage{multicol}
\usepackage[table]{xcolor}
\usepackage{url}

\newcommand{\PreserveBackslash}[1]{\let\temp=\\#1\let\\=\temp}
\newcolumntype{C}[1]{>{\PreserveBackslash\centering}p{#1}}
\newcolumntype{R}[1]{>{\PreserveBackslash\raggedleft}p{#1}}
\newcolumntype{L}[1]{>{\PreserveBackslash\raggedright}p{#1}}
\graphicspath{{figure/}}

\usepackage{algorithm}
\usepackage{algorithmic}

\usepackage{flushend}
\usepackage{ragged2e}

\begin{document}

\title{Weighted Dark Channel Dehazing}

\author{Mingzhu~Zhu, Bingwei~He, Jiantao~Liu
\IEEEcompsocitemizethanks{\IEEEcompsocthanksitem
The authors were
with the Department of Mechanical Engineering of Fuzhou University, Fuzhou,
FuJian, China.
E-mail: mebwhe@fzu.edu.cn
}
\thanks{This work was supported by the National Natural Science Foundation
of China (Project No. 61473090 and No.61673115) and Fujian Provincial
Collaborative Innovation Center for High-End Equipment Manufacturing.}}

\markboth{}%
{Shell \MakeLowercase{\textit{et al.}}: Bare Demo of IEEEtran.cls for Computer Society Journals}

\IEEEtitleabstractindextext{%
\begin{abstract}
\justifying
In dark channel based methods, local constant assumption is widely used to make the algorithms invertible.
It inevitably introduces defects since the assumption can not perfectly avoid depth discontinuities and meanwhile cover enough pixels.
Unfortunately, because of the limitation of the prior, which only confirms the existence of dark things but does not specify their locations or likelihood, no fidelity measurement is available in refinement thus the defects are either under-corrected or over-corrected.
In this paper, we go deeper than the dark channel theory to overcome this problem.
We split the concept of dark channel into dark pixels and local constant assumption,
and then,
control the problematic assumption based on a novel weight map.
With such effort, our methods show significant improvement on quality and have competitive speed.
In the last, we show that the method is highly robust to initial transmission estimates and can be ever-improved by providing better dark pixel locations.
\end{abstract}

}
\maketitle
\IEEEdisplaynontitleabstractindextext
\IEEEpeerreviewmaketitle

\IEEEraisesectionheading{\section{Introduction}\label{sec:introduction}}
\IEEEPARstart{H}{azy} scenes widely exist in outdoor environment.
The observed images suffer from low-contrast and visibility.
Most of the computer vision applications assume that input image is scene radiance, therefore, haze removal is highly desired, whether by enhancing hazy image \cite{ancuti2013single,galdran2017fusion,guo2017efficient} or recovering haze-free image \cite{he2011single,fattal2014dehazing,berman2016non}.

The model widely used to describe hazy image is~\cite{middleton1957vision}
\begin{equation}
\label{Eq:hazy model}
I(x)=t(x)J(x)+[1-t(x)]A,
\end{equation}
where $x$ is the pixel coordinate, $I$ is the observed image, $J$ is the scene radiance, $A$ is the global atmospheric color, which can be estimated by selecting particular pixels \cite{he2011single}, or statistical approaches \cite{bahat2016blind,berman2017air}.
In this model, $I$ is defined as a fusion of $J$ and $A$ with the ratio controlled by transmission map $t$, which depends on attenuation coefficient $\beta$ and scene depth $d$,
as $t(x)=e^{-\beta d(x)}$.
Commonly, $\beta$ is assumed to be constant, thus $t$ is inversely correlated to $d$.

The recovery of scene radiance is an under-constrained problem.
Earlier methods depend on additional inputs \cite{Narasimhan2000Chromatic,schechner2001instant,kopf2008deep}, therefore, limiting the applicability.
Researchers have developed various assumptions, models and priors \cite{he2011single,fattal2014dehazing,berman2016non} for the more challenging single-input cases.
Among them, the dark channel prior~\cite{he2011single} is widely recognized, which is defined as
\begin{equation}
\label{Eq:dc prior}
\min_{c}(\min_{y\in\Omega(x)}J^c(y))=0,
\end{equation}
where $c$ is the color channel, and $\Omega(x)$ is the mask centered at $x$.
Despite the prior describes haze-free images in high quality, it does not provide sufficient constraints.
Relevant methods have to employ extra assumptions to make the problem invertible.
However, these assumptions are usually not rigorous enough, resulting in various defects.

\begin{figure}[t]
\centering
\begin{subfigure}{0.42\linewidth}\includegraphics[width=\textwidth,height=0.6\textwidth]{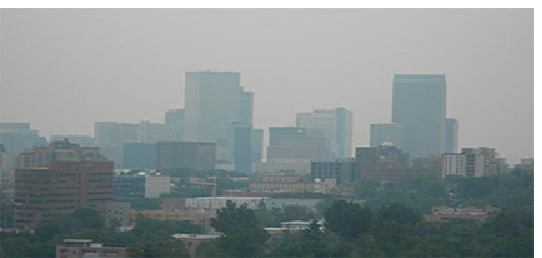}
\caption{hazy image}
\end{subfigure}
\begin{subfigure}{0.42\linewidth}\includegraphics[width=\textwidth,height=0.6\textwidth]{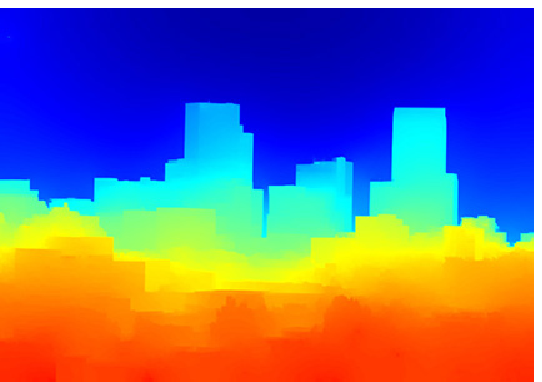}
\caption{trans. of our method}
\label{subfig:DCshow-iniWDC}
\end{subfigure}
\\
\begin{subfigure}{0.42\linewidth}\includegraphics[width=\textwidth,height=0.6\textwidth]{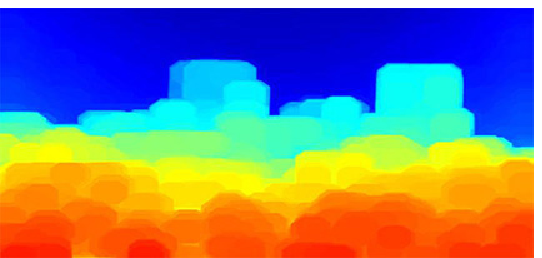}
\caption{initial trans. of~\cite{he2011single}}
\label{subfig:DCshow-iniHe}
\end{subfigure}
\begin{subfigure}{0.42\linewidth}\includegraphics[width=\textwidth,height=0.6\textwidth]{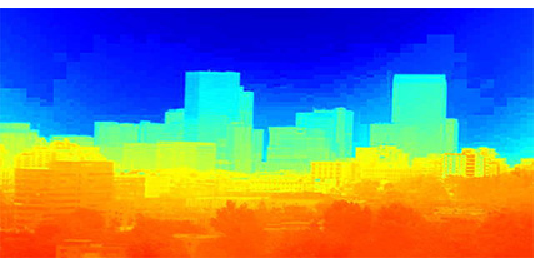}
\caption{trans. of~\cite{he2011single}}
\label{subfig:DCshow-He}
\end{subfigure}
\\
\begin{subfigure}{0.42\linewidth}\includegraphics[width=\textwidth,height=0.6\textwidth]{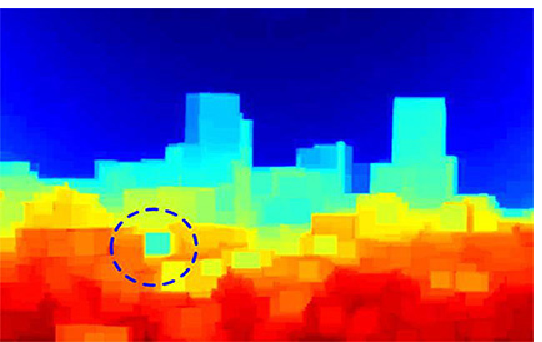}
\caption{initial trans. of~\cite{meng2013efficient}}
\label{subfig:DCshow-iniMeng}
\end{subfigure}
\begin{subfigure}{0.42\linewidth}\includegraphics[width=\textwidth,height=0.6\textwidth]{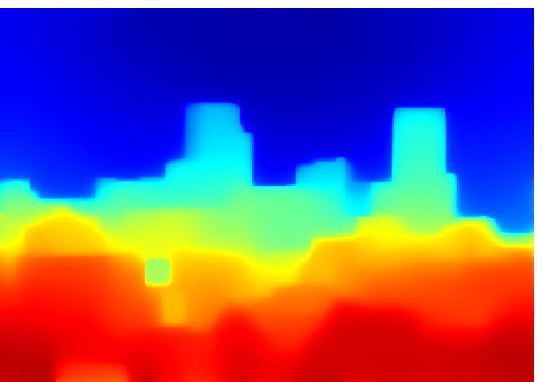}
\caption{trans. of~\cite{meng2013efficient}}
\label{subfig:DCshow-Meng}
\end{subfigure}
\\
\caption{\textbf{Intermediates of several dark channel based methods.}
Warmer color indicates higher transmission.
Note that, transmission map should well and only reflect scene depth.}
\label{fig:DCshow}
\end{figure}

He~\emph{et~al.}~\cite{he2011single} assumes that transmissions are constant in each $\Omega$, thus $t(y)=\tilde{t}(x)$ for $y\in\Omega(x)$,
where $\tilde{t}(x)$ is the transmission of the patch centered at $x$.
Combining this assumption and the prior, the transmissions are initially estimated by dilating balanced input, as
\begin{equation}
t=1-\min_{c}(\min_{y\in\Omega(x)}\frac{I^c(y)}{A^c}).
\end{equation}
The estimates have block effect as shown in Fig.~\ref{subfig:DCshow-iniHe}, where transmissions in the vicinity of depth discontinuities are mostly over-estimated.
Laplacian matting~\cite{levin2008closed} is then employed to solve this problem with local linear assumption between transmissions and hazy colors,
which is not suitable in all cases.
The refined estimates mix depth discontinuities and color textures, containing redundant details as shown in Fig.~\ref{subfig:DCshow-He}.
This refinement will lead to micro-contrast loss in the result~\cite{fattal2014dehazing,chen2016robust,he2017haze}.
Furthermore, this method is relatively slow due to the Laplacian matting.
Although it can be replaced by guided filter~\cite{he2013guided} for efficiency, severe halo-effect will be introduced.

As an improvement,
Meng~\emph{et~al.}~\cite{meng2013efficient} starts from applying image opening instead of dilation in initial transmission estimation, as (suppose the radiance bounds are 0 and 1)
\begin{equation}
\label{Eq:Meng initial}
t=1-\max_{y\in\Omega(x)}(\min_{c,z\in\Omega(y)}\frac{I^c(z)}{A^c}).
\end{equation}
An example is shown in Fig.~\ref{subfig:DCshow-iniMeng}.
Although the edges are more in accordance with depth discontinuities compared to Fig.~\ref{subfig:DCshow-iniHe}, some ill-defined boundaries appear.
From the perspective of dark channel prior, Eq.~(\ref{Eq:Meng initial}) uses local constant assumption on space-variant mask, which degrades the accuracy of the prior, resulting in failures such as the circled region in Fig.~\ref{subfig:DCshow-iniMeng}.
The estimates are refined by minimizing a cost function based on piecewise smoothness assumption, which is more reasonable than the local linear assumption~\cite{chen2016robust}.
However, the optimal solution is searched by an algorithm with initial sensitive problem.
Edges not included in initial estimates will never appear and wrong estimated edges can only be smoothed.
Consequently, the final estimates are over-smoothed as shown in Fig.~\ref{subfig:DCshow-Meng}.

Wang~\emph{et~al.}~\cite{wang2016haze} uses the local constant assumption on super-pixels.
It relies on the quality of super-pixel segmentation.
Block effect will appear if the super-pixels are too large to avoid depth discontinuities, while the prior will be undermined if the super-pixels are too small to cover white objects.
Guided filter~\cite{he2013guided} is employed to improve the stability, but causes halo-effect.
Similarly, the adaptive mask proposed by Fang~\emph{et~al.}~\cite{fang2014single} undermines the prior since the masks only cover pixels with similar colors.

\begin{figure*}[t]
\centering
\includegraphics[width=6in]{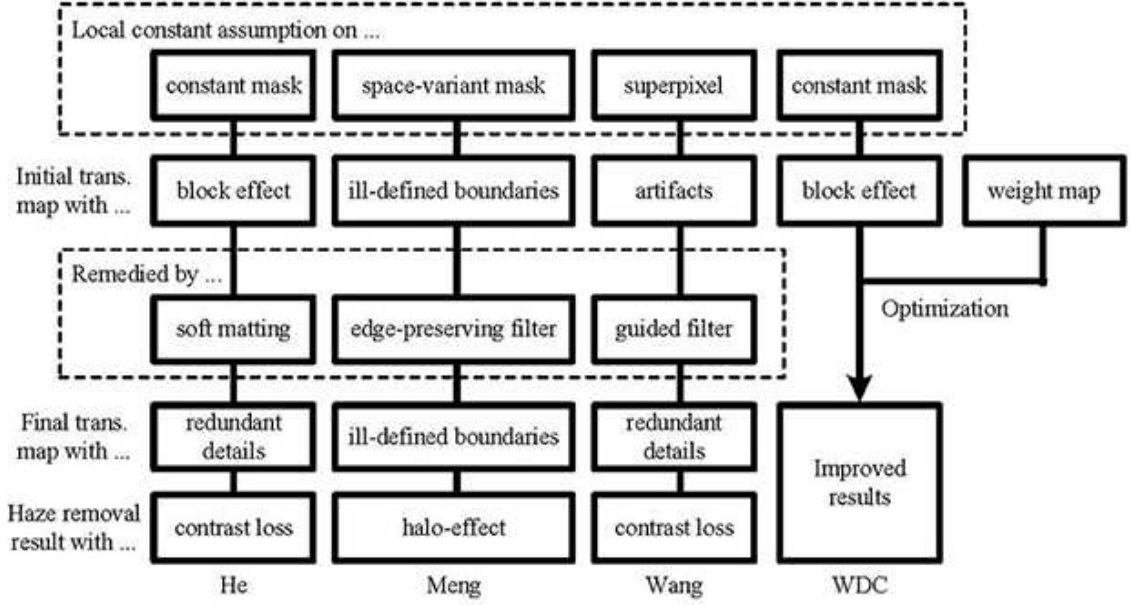}
\caption{\textbf{Flowcharts of several dark channel prior based methods.}
He~\emph{et~al.}~\cite{he2011single},
Meng~\emph{et~al.}~\cite{meng2013efficient},
Wang~\emph{et~al.}~\cite{wang2016haze},
and WDC.
Developing a fidelity measurement could fundamentally solve various problems in dark channel based methods.}
\label{fig:Flowcharts}
\end{figure*}

Several methods above are summarized in Fig.~\ref{fig:Flowcharts}.
It shows that improper mask shapes are the roots of their defects.
Ill-defined initial transmission estimates are inevitable since the masks (or segmentations) can not be perfect.
Most researches focus on post-refinement,
however, the local linear assumption based algorithms will introduce redundant details~\cite{li2015edge,zhu2015fast,cai2016dehazenet},
and the piecewise smoothness assumption based algorithms will lead to over-smoothness if fidelity measurement is absent~\cite{He2016Convex}.
Many researches develop edge-preserving filters~\cite{lai2015single,chen2016robust}, but end up in preserving strong edges rather than depth discontinuities.

Instead of offering constraints on local regions,
some recently proposed priors offer pixel-level constraints directly.
For example, Fattal~\cite{fattal2014dehazing} fits color-lines in RGB space per-patch, which are supposed to cross the origin in haze-free images.
These lines are shifted in the direction of atmospheric color $A$ by haze, thus the transmissions of the pixels on them could be estimated from their $A$-intercept.
However, the model is incomplete because there are also many color-lines that do not cross the origin in haze-free images~\cite{omer2004color}.
Bahat and Irani~\cite{bahat2016blind} estimates transmission map by maximizing the number of internal patch recurrence.
Although haze-free images usually have more recurrent patches, surplus textures will appear when the number is simply maximized.
Berman~\emph{et~al.}~\cite{berman2016non} gathers pixels that are supposed to have the same radiance as a haze-line.
On each haze-line, the outermost pixel is assumed to be haze-free, thus the transmissions of other pixels could be deduced.
However, haze-line model will fail when the input lacks of color, and the haze-free assumption will fail when the input lacks of haze-free regions, which is a common case.

Despite being flawed in prior, these methods have high quality results in some cases.
A common advantage is that their initial transmission estimates contain fidelity measurements.
In Fattal~\cite{fattal2014dehazing}, the weight of a patch depends on the colinearity of the pixels within.
In Bahat and Irani~\cite{bahat2016blind}, the weight of a pixel is correlated to its significance on the recurrent patches number.
In Berman~\emph{et~al.}~\cite{berman2016non}, the weight of a haze-line depend on its effective length.
Therefore, their initial estimates of the pixels which are not well described by their priors affect the results little.

Learning-based methods are out of the framework above.
Some networks only provide partially reliable transmissions, thus post-refinement is still required.
For example, Cai~\emph{et~al.}~\cite{cai2016dehazenet}, Song~\emph{et~al.}~\cite{song2018single} and Li~\emph{et~al.}~\cite{li2018cascaded} employ guided filter to refine their raw transmissions, leading to the same problems.
As improvements, Zhang and Patel~\cite{zhang2018densely} avoids over-smoothness by using an edge-preserving loss function in training.
Its transmission maps are inherently regularized.
Li~\emph{et~al.}~\cite{Li2017AOD} and Li~\emph{et~al.}~\cite{li2018single} propose end-to-end dehazing networks which do not explicitly produce transmission map.
However, different from the ever-improving training strategy, there is little improvement on building training set.
Due to the lack of real-scene haze-free and hazy image pairs, training set are built from synthesized images.
Indoor or close-shot images are commonly used as haze-free images for available depth maps.
The mismatch between synthesized training set and real-scene image pairs might lead to frequent bad cases.
Fan~\emph{et~al.}~\cite{fan2017two} use real-world hazy images but the corresponding transmission maps are generated by traditional methods~\cite{fattal2008single,he2011single,Tang2014Investigating},
which might inherit the defects of these traditional methods.

In summary, the dark channel prior based methods are below the full potential of their prior because the absent of fidelity measurement for local constant assumption.
In this paper, WDC (weighted dark channel dehazing) is proposed.
As illustrated in Fig.~\ref{fig:Flowcharts}, a weight map is introduced to avoid these common defects.
Based on WDC, two extensions are proposed.
The first one named CWDC (constrained WDC) provides transmission estimates in a higher quality by considering transmission lower bound exactly.
The second extension named EWDC (extensible WDC) provides WDC with an interface which receives extra messages provided by manual operations or other theories to tune the results.

The motivation of fully exploiting the power of the dark channel prior is inherited from our previous work~\cite{Zhu2017Single},
whose major cost function is over-complicate and produces transmissions in discrete space (5-bit) due to the introduction of label set at every pixel.
WDC is much faster, more concise and works in continuous space.
Furthermore, the output quality is further improved by CWDC.

The remainder of this paper is organized as follows.
WDC is introduced in Section~\ref{sec:WDC}.
CWDC is introduced in Section~\ref{sec:CWDC}.
The experiments are displayed in Section~\ref{sec:EXP}.
Because of the introduction of extra messages, EWDC and its two instances are not included in the comparison, but illustrated in Section~\ref{sec:EWDC}.
The conclusions are given in Section~\ref{sec:CON}.

\section{WDC}
\label{sec:WDC}

\begin{algorithm}[h]
\caption{WDC}
\label{Alg:dehaze}
\begin{algorithmic}[1]
\REQUIRE    $I,A$
\ENSURE     $J,t$
\STATE lower bound $b(x)=1-\min_c\frac{I^c(x)}{A^c}$
\STATE initial trans. map $\tilde{t}(x)=\max_{y\in\Omega(x)}b(y)$
\STATE weight map $W(x)=1/(\tilde{t}(x)-b(x))^2$
\STATE calculate $t$ by Eq.~(\ref{Eq:refinement-matrix-solution})
\STATE calculate $J$ by Eq.~(\ref{Eq:dehaze})
\end{algorithmic}
\end{algorithm}

\begin{figure}[t]
\centering
\begin{subfigure}{0.4\linewidth}\includegraphics[width=\textwidth,height=0.6\textwidth]{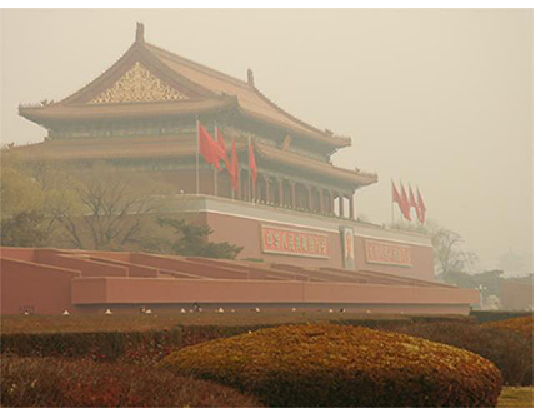}
\caption{$I$}
\end{subfigure}
\begin{subfigure}{0.4\linewidth}\includegraphics[width=\textwidth,height=0.6\textwidth]{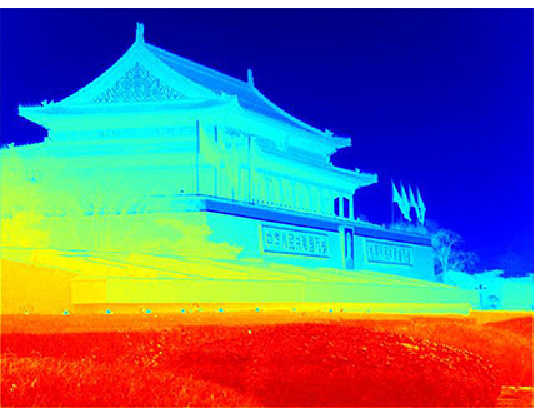}
\caption{$b$}
\end{subfigure}
\\
\centering
\begin{subfigure}{0.4\linewidth}\includegraphics[width=\textwidth,height=0.6\textwidth]{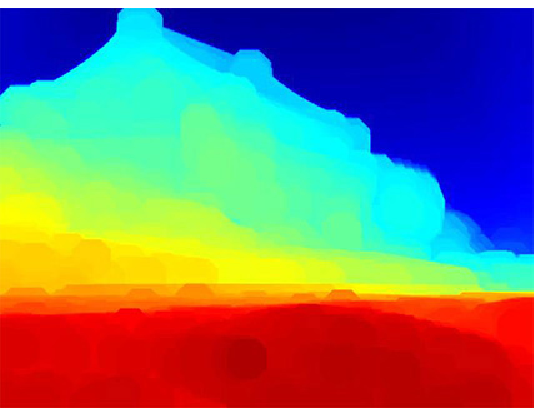}
\caption{$\tilde{t}$}
\label{subfig:Flow-ini}
\end{subfigure}
\begin{subfigure}{0.4\linewidth}\includegraphics[width=\textwidth,height=0.6\textwidth]{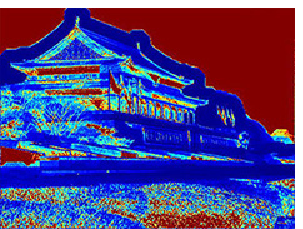}
\caption{$W$}
\label{fig:Flow-weight}
\end{subfigure}
\\
\centering
\begin{subfigure}{0.4\linewidth}\includegraphics[width=\textwidth,height=0.6\textwidth]{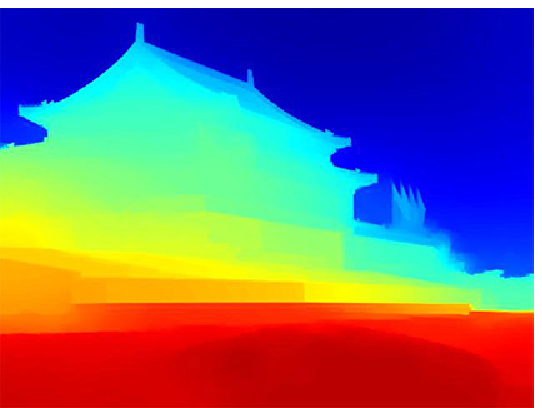}
\caption{$t$}
\end{subfigure}
\begin{subfigure}{0.4\linewidth}\includegraphics[width=\textwidth,height=0.6\textwidth]{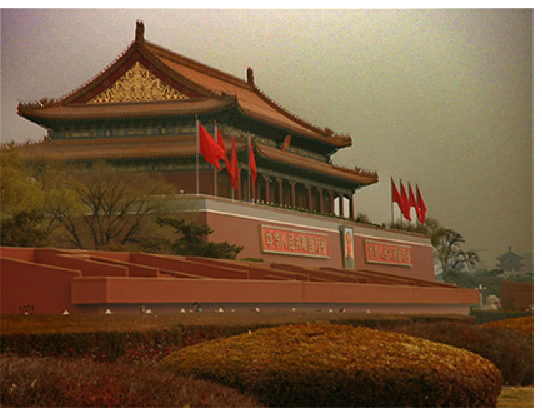}
\caption{$J$}
\end{subfigure}
\\
\caption{\textbf{Intermediate results of WDC.}}
\label{fig:Flow}
\end{figure}

\begin{figure*}[t]
\centering
\begin{subfigure}{0.22\linewidth}\includegraphics[width=\textwidth,height=0.6\textwidth]{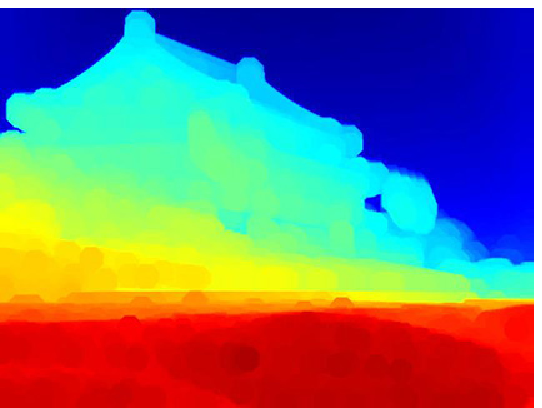}
\end{subfigure}
\begin{subfigure}{0.22\linewidth}\includegraphics[width=\textwidth,height=0.6\textwidth]{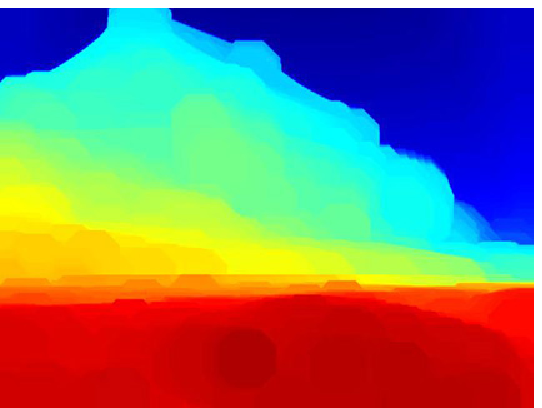}
\end{subfigure}
\begin{subfigure}{0.22\linewidth}\includegraphics[width=\textwidth,height=0.6\textwidth]{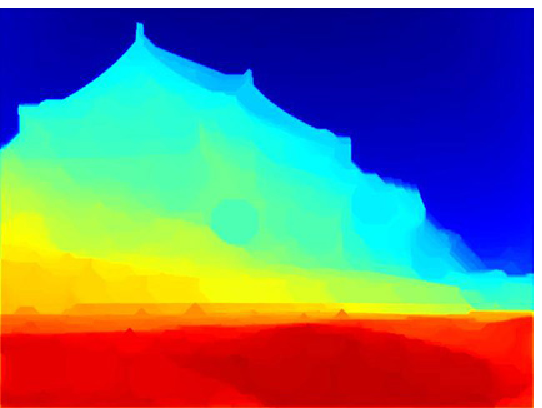}
\end{subfigure}
\begin{subfigure}{0.22\linewidth}\includegraphics[width=\textwidth,height=0.6\textwidth]{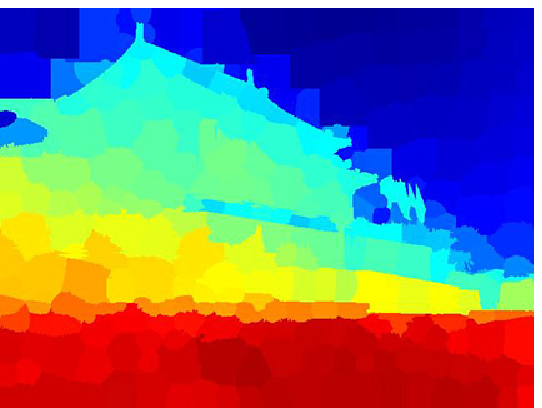}
\end{subfigure}
\\
\centering
\begin{subfigure}{0.22\linewidth}\includegraphics[width=\textwidth,height=0.6\textwidth]{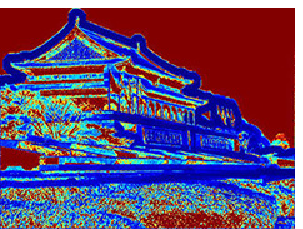}
\end{subfigure}
\begin{subfigure}{0.22\linewidth}\includegraphics[width=\textwidth,height=0.6\textwidth]{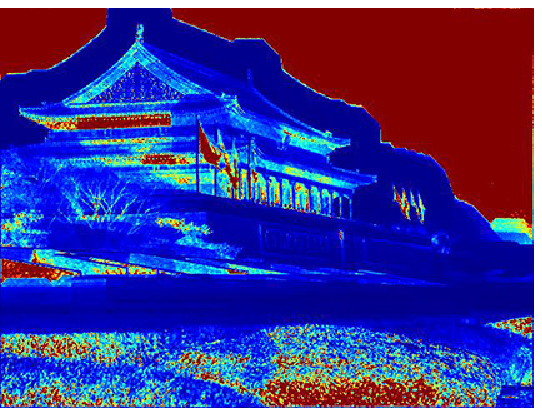}
\end{subfigure}
\begin{subfigure}{0.22\linewidth}\includegraphics[width=\textwidth,height=0.6\textwidth]{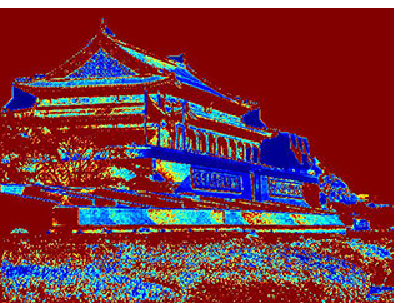}
\end{subfigure}
\begin{subfigure}{0.22\linewidth}\includegraphics[width=\textwidth,height=0.6\textwidth]{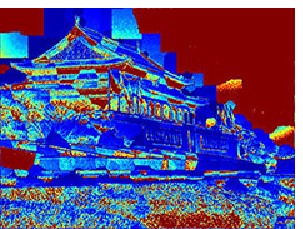}
\end{subfigure}
\\
\centering
\begin{subfigure}{0.22\linewidth}\includegraphics[width=\textwidth,height=0.6\textwidth]{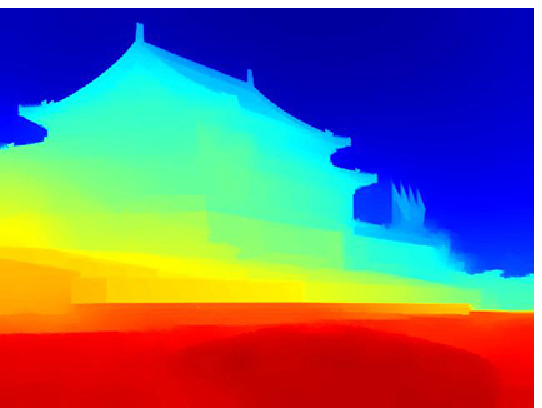}
\caption{$r(\Omega)=15$}
\end{subfigure}
\begin{subfigure}{0.22\linewidth}\includegraphics[width=\textwidth,height=0.6\textwidth]{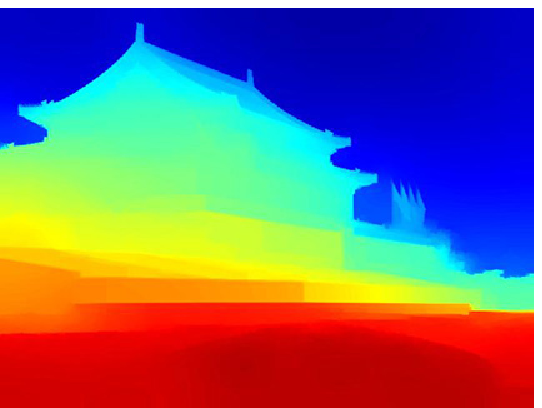}
\caption{$r(\Omega)=35$}
\end{subfigure}
\begin{subfigure}{0.22\linewidth}\includegraphics[width=\textwidth,height=0.6\textwidth]{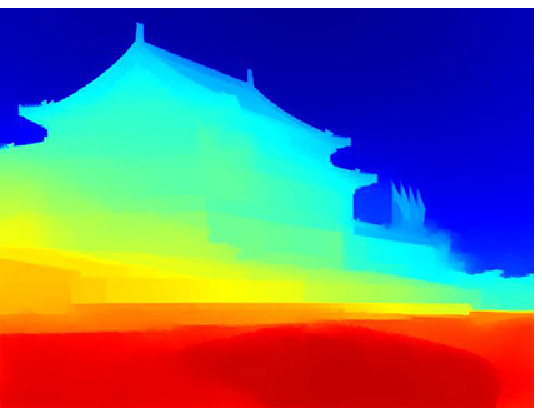}
\caption{image opening~\cite{meng2013efficient}}
\end{subfigure}
\begin{subfigure}{0.22\linewidth}\includegraphics[width=\textwidth,height=0.6\textwidth]{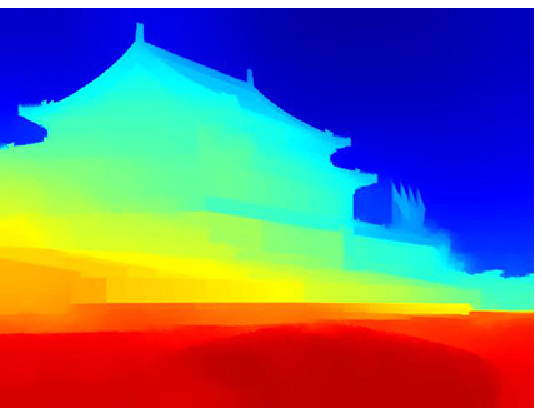}
\caption{super-pixel~\cite{wang2016haze}}
\end{subfigure}
\\
\caption{\textbf{Intermediate results of WDC with different local constant assumptions,
showing that WDC is robust to initial estimates.}
From top to bottom, initial trans. map $\tilde{t}$, weight map $W$ and final trans. map $t$.}
\label{fig:Robust}
\end{figure*}

According to Eq.~(\ref{Eq:hazy model}), $t(x)$ can be expressed as
\begin{equation}
t(x)=\frac{1-\min_c(I^c(x)/A^c)}{1-\min_c(J^c(x)/A^c)}.
\end{equation}
Since $J^c(x)/A^c$ is non-negative, a lower bound of $t$ is
\begin{equation}
t(x)\ge{b(x)}=1-\min_{c}({I^c(x)}/{A^c}),
\end{equation}
where $A$ is estimated using previous methods~\cite{berman2017air,he2011single}.
For pixels $z$ satisfying $\min_cJ^c(z)=0$ (named dark pixel), we get
\begin{equation}
\label{Eq:dark pixel}
t(z)=1-\min_{c}({I^c(z)}/{A^c})=b(z),
\end{equation}
which means that the transmissions of dark pixels equal to their lower bounds.

If we use the local constant assumption on the $\Omega(x)$ defined in Eq.~(\ref{Eq:dc prior}), and denote the transmission of $\Omega(x)$ as $\tilde{t}(x)$, then $t(y)=\tilde{t}(x)$ for $y\in\Omega(x)$.
Because of the lower bound constraint, $\tilde{t}(x)$ must be no less than any $b(y)$, as
\begin{equation}
\label{Eq:lower bound}
\tilde{t}(x)\ge\max_{y\in\Omega(x)}b(y)=1-\min_{y\in\Omega(x)}(\min_c{({I^c(x)}/{A^c}})).
\end{equation}
Denote $Z(x)$ as the set of dark pixels in $\Omega(x)$.
The dark channel prior indicates that $Z(x)$ contains at least one dark pixel.
For each $z\in{Z}(x)$, we have
\begin{equation}
t(z)=b(z)=\tilde{t}(x)\ge\max_{y\in\Omega(x)}b(y),
\end{equation}
which is based on Eq.~(\ref{Eq:dark pixel}), the local constant assumption and Eq.~(\ref{Eq:lower bound}) respectively.
Since $b(z)\ge\max_{y\in\Omega(x)}b(y)$ and $Z(x)$ is a subset of $\Omega(x)$, we have $b(z)=\max_{y\in\Omega(x)}b(y)$, thus
\begin{equation}
\label{Eq:constant result}
\tilde{t}(x)=\max_{y\in\Omega(x)}b(y).
\end{equation}

As shown above, the local constant assumption on $\Omega$ leads to two deductions.
Firstly, transmission map is the $\tilde{t}$ of Eq.~(\ref{Eq:constant result}).
Secondly, pixels satisfying $b(x)=\tilde{t}(x)$, namely the local maximum, are dark pixels.
The local constant assumption is false at depth discontinuities, where the first deduction will fail and lead to block effect.
However, the second deduction is relatively robust.
As shown in Fig.~\ref{fig:Robust}, despite the different masks, most of the likely dark pixels (red pixels in the second row) appear in the right positions, including those of shadowed, colorful and dark things.
Therefore, we use it to measure the fidelity of $\tilde{t}$.

At each pixel, $\tilde{t}(x)$ suggests $t(x)=b(z)$, where $z=\arg\max_{y\in\Omega(x)}b(y)$.
To measure the fidelity of $t(x)=b(z)$, we firstly estimate the likelihood of $z$ being a dark pixel by
\begin{equation}
\label{Eq:weight map-d}
W_d(z)=f(|\tilde{t}(z)-b(z)|),
\end{equation}
where $f$ is a decreasing function.
Then, suppose $z$ is a dark pixel, namely $t(z)=b(z)$, the fidelity of $t(x)=b(z)$ equals to the fidelity of $t(x)=t(z)$, which can be measured by $1/(I(x)-I(z))^2$.
We use $1/(b(x)-b(z))^2$ instead for two reasons.
Firstly, pixels similar in color are also similar in $b$.
Secondly, $b(z)$ equals $\tilde{t}(x)$ thus we do not need to store the position of $z$.
Therefore, the overall fidelity of $\tilde{t}(x)$ is
\begin{equation}
\label{Eq:weight map}
W(x)=W_d(z)W_s(x)\approx{kW_s(x)}={k}/{(\tilde{t}(x)-b(x))^2},
\end{equation}
where $k$ is a constant.
The approximation of $W_d=k$ is based on the observation that $\tilde{t}(z)$ and $b(z)$ are usually the same.
In practise, we set a lower bound for $\tilde{t}(x)-b(x)$ as $10^{-3}$ and normalize $W$, thus $k$ is actually not required.

The initial transmission map $\tilde{t}$ and its weight map $W$
are in the data term of our cost function, which is
\begin{equation}
\label{Eq:refinement}
\left\{
\begin{aligned}
E(t)&=\sum_x{W(x)(t(x)-\tilde{t}(x))^2}+\lambda\sum_{(x,y)\in{N}}\frac{(t(x)-t(y))^2}{||I(x)-I(y)||^2}\\
t&\geq{b}\\
\end{aligned}
\right.
\end{equation}
where $(x,y)\in{N}$ means that $x$ and $y$ are adjacent pixels,
and $\lambda$ controls the degree of smoothness ($0.02$ in this paper).
This is a huge-scale constrained QP problem.
Searching the optimal solution is time-consuming, thus we ignore the constraint like other methods~\cite{fattal2014dehazing,berman2016non,chen2016robust}, and solve $t$ as
\begin{equation}
\label{Eq:refinement-matrix-solution}
t=(W+\lambda{L})^{-1}W\tilde{t},
\end{equation}
where $W$ is in matrix form and $L$ is a Laplacian matrix.

With $A$ and $t$, the haze removal result is given by
\begin{equation}
\label{Eq:dehaze}
J(x)=\frac{I(x)-A}{(\max(t(x),b)+\epsilon_t)/(1+\epsilon_t)}+A,
\end{equation}
where $t$ is slightly increased by $\epsilon_t$ to suppress the noise at far distance and to compensate the residual errors, which are from the prior and the ignoring of lower bound constraint.

WDC is summarized in Alg.~\ref{Alg:dehaze} and demonstrated in Fig.~\ref{fig:Flow}.
The intuition of the weight map can be better understood by checking the following points.

\begin{itemize}
\item Large $W$ only appear on likely dark pixels, which have $t(x)\approx\tilde{t}(x)$.
    The estimates of these high fidelity pixels are propagated to the surrounding pixels in refinement.
    As shown in Fig.~\ref{subfig:Flow-ini}, the transmissions of the sky region in the vicinity of the building are over-estimated.
    However, their fidelities are very low (with small values in Fig.~\ref{fig:Flow-weight}), thus their final estimates $t$ are propagated from other parts of the sky.
\item Low $W(x)$ does not mean $t(x)$ should be different from $\tilde{t}(x)$, but means $t(x)$ should refer more to its neighbors.
\item The data term of our cost function is based on the likely dark pixels rather than the whole initial estimates.
    As a benefit, the final estimates are robust to the shapes of masks, as shown in the last row of Fig.~\ref{fig:Robust}.
\end{itemize}

\section{CWDC}
\label{sec:CWDC}

\begin{figure}[t]
\centering
\begin{subfigure}{0.4\linewidth}\includegraphics[width=\textwidth,height=0.6\textwidth]{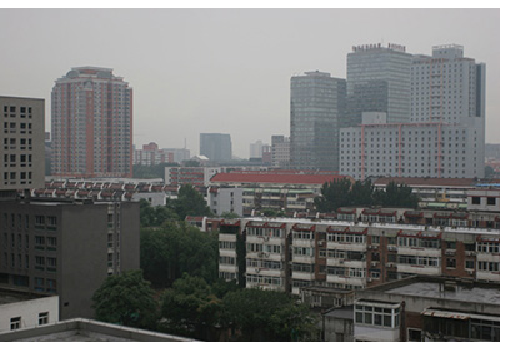}
\caption{$I_1$}
\end{subfigure}
\begin{subfigure}{0.4\linewidth}\includegraphics[width=\textwidth,height=0.6\textwidth]{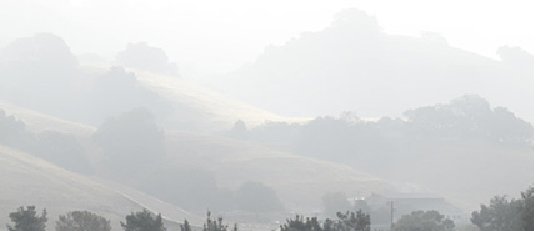}
\caption{$I_2$}
\label{subfig:WhyCWDC-b}
\end{subfigure}
\\
\centering
\begin{subfigure}{0.4\linewidth}\includegraphics[width=\textwidth,height=0.6\textwidth]{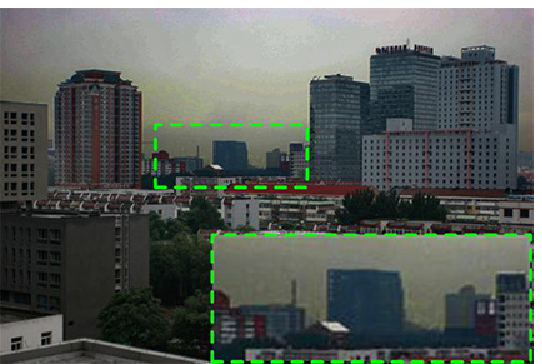}
\caption{$J_{wdc1}$ with $\epsilon_t$=0.1}
\label{subfig:WhyCWDC-c}
\end{subfigure}
\begin{subfigure}{0.4\linewidth}\includegraphics[width=\textwidth,height=0.6\textwidth]{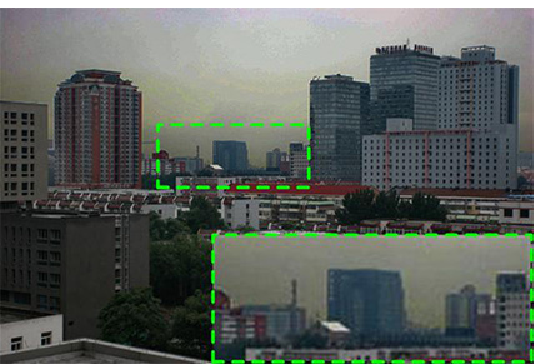}
\caption{$J_{cwdc1}$ with $\epsilon_t$=0.1}
\label{subfig:WhyCWDC-d}
\end{subfigure}
\\
\centering
\begin{subfigure}{0.4\linewidth}\includegraphics[width=\textwidth,height=0.6\textwidth]{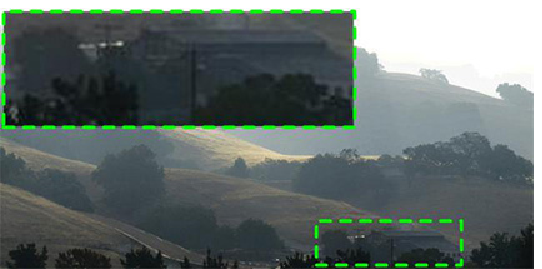}
\caption{$J_{wdc2}$ with $\epsilon_t$=0.1}
\label{subfig:WhyCWDC-e}
\end{subfigure}
\begin{subfigure}{0.4\linewidth}\includegraphics[width=\textwidth,height=0.6\textwidth]{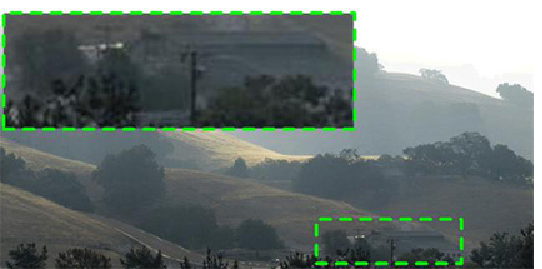}
\caption{$J_{cwdc2}$ with $\epsilon_t$=0.1}
\label{subfig:WhyCWDC-f}
\end{subfigure}
\\
\centering
\begin{subfigure}{0.4\linewidth}\includegraphics[width=\textwidth,height=0.6\textwidth]{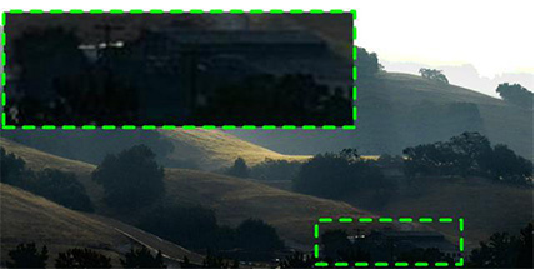}
\caption{$J_{wdc2}$ with $\epsilon_t$=0.01}
\label{subfig:WhyCWDC-g}
\end{subfigure}
\begin{subfigure}{0.4\linewidth}\includegraphics[width=\textwidth,height=0.6\textwidth]{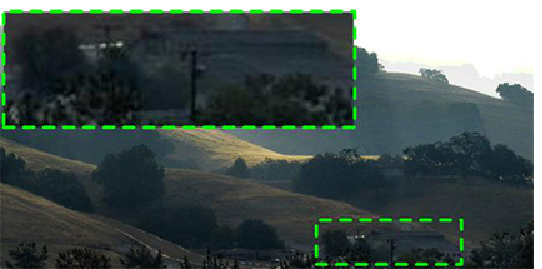}
\caption{$J_{cwdc2}$ with $\epsilon_t$=0.01}
\label{subfig:WhyCWDC-h}
\end{subfigure}
\\
\caption{\textbf{Comparison of WDC and CWDC.}}
\label{fig:WhyCWDC}
\end{figure}

WDC ignores the lower bound of Eq.~(\ref{Eq:refinement}), thus the optimal solution might be smaller than $b$, leading to negative values in $J$.
This approximation also exists in other optimization based methods~\cite{fattal2014dehazing,berman2016non,chen2016robust}
because traditional algorithms such as interior point method or gradient projection method can not solve this huge-scale constrained optimization problem in an acceptable runtime.
The approximate solutions are usually compensated by two tricks.
The first one is applying $t=max(t,b)$ to prevent negative result.
The second one is increasing the haze preservation parameter (the $\epsilon_t$ of Eq.~(\ref{Eq:dehaze}) in this paper).

In this section, we extend WDC by taking the low-bound constraint into account.
We modify Eq.~(\ref{Eq:refinement}) by optimizing the gap between $b$ and $t$ instead, as
\begin{equation}
\label{Eq.CWDC-energy}
E(x)=\frac{1}{2}x^TQx+c^Tx,\quad x\ge0,
\end{equation}
where $x=t-b$, $Q=2W+2\lambda{L}$, $c=2(b-\tilde{t})^TW+2b^TL$.
Its optimal solution can be found within a much more reasonable time based on Fletcher~\cite{Fletcher2017Augmented},
which to our best knowledge is the fastest non-negative QP algorithm.
Readers could refer to \cite{Rockafellar1974Augmented,Fletcher2017Augmented} for convergence proof and other details.

With closely satisfied lower bound constraint, our method reaches its full potential, denoted as CWDC.
Fig.~\ref{fig:WhyCWDC} shows the comparison of WDC and CWDC.
The situation of $t<b$ usually happens in blocks,
where the trick of $t=max(t,b)$ results in $t=b$.
It mixes color textures with depth discontinuities, resulting in micro-contrast loss, as shown in the zoomed-in region of Fig.~\ref{subfig:WhyCWDC-c}.
As comparison, details of these distant buildings are clear in Fig.~\ref{subfig:WhyCWDC-d}.
The trick of compensation parameter ($\epsilon_t$ in this paper) needs to find a balance between haze removal and negative result prevention.
To achieve the best result, $\epsilon_t$ should varies with image contents and sometimes does not exist.
As shown in Fig.~\ref{subfig:WhyCWDC-e},
$\epsilon_t=0.1$ is not small enough to remove haze in the distance,
but not big enough to avoid negative results, where the structures of the trees are blurred.
For the case of $\epsilon_t=0.01$, haze is well removed but the problem of negative result is worse.
In both configurations, CWDC preserves details well, which is most obvious on the trees.

\section{Experiments}
\label{sec:EXP}

\begin{figure}[t]
\centering
\begin{subfigure}{0.40\linewidth}\includegraphics[width=\textwidth,height=0.6\textwidth]{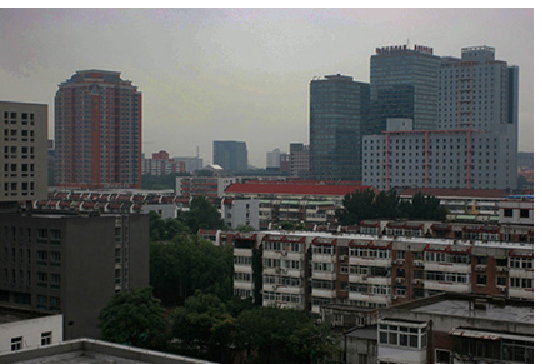}
\end{subfigure}
\begin{subfigure}{0.40\linewidth}\includegraphics[width=\textwidth,height=0.6\textwidth]{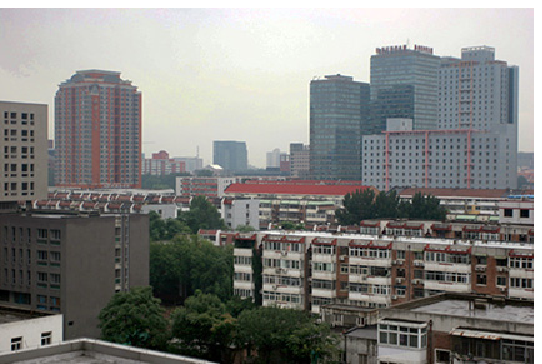}
\end{subfigure}
\\
\centering
\begin{subfigure}{0.40\linewidth}\includegraphics[width=\textwidth,height=0.6\textwidth]{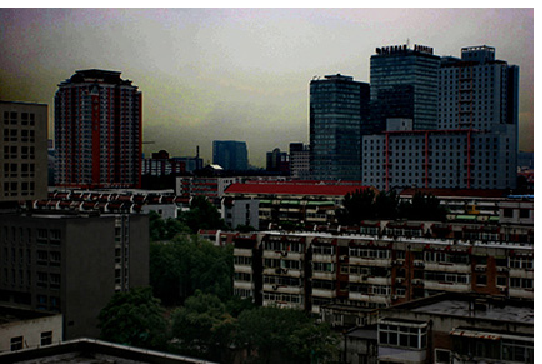}
\end{subfigure}
\begin{subfigure}{0.40\linewidth}\includegraphics[width=\textwidth,height=0.6\textwidth]{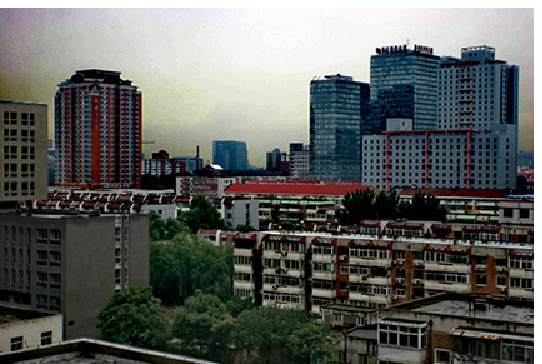}
\end{subfigure}
\\
\centering
\begin{subfigure}{0.40\linewidth}\includegraphics[width=\textwidth,height=0.6\textwidth]{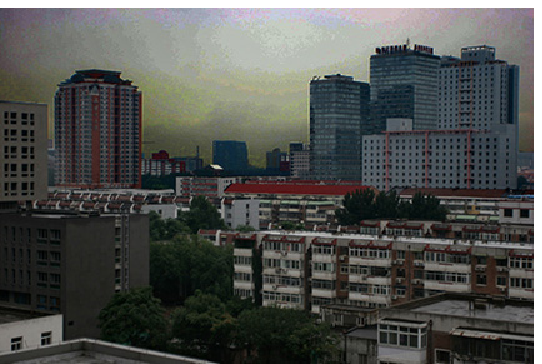}
\caption{without post-process}
\end{subfigure}
\begin{subfigure}{0.40\linewidth}\includegraphics[width=\textwidth,height=0.6\textwidth]{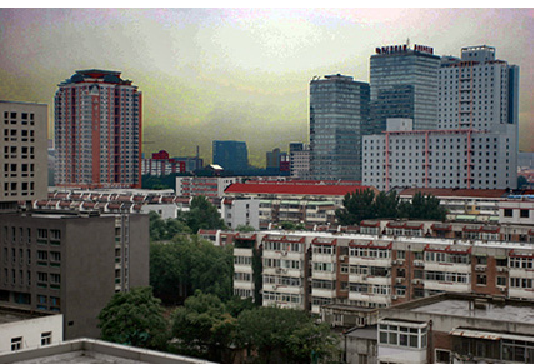}
\caption{with post-process}
\end{subfigure}
\\
\caption{\textbf{Comparison of the results with and without post-enhancement (contrast-stretch~\cite{berman2016non}).}
From top to bottom, He~\emph{et~al.}~\cite{he2011single},
Berman~\emph{et~al.}~\cite{berman2016non},
and WDC.}
\label{fig:Post}
\end{figure}

\begin{table*}[t]
\centering
\caption{\textbf{Comparison on D-HAZY dataset~\cite{ancuti2016d}.}}
\label{Table:D-HAZY}
\begin{tabular}{C{1.2cm}|C{1.2cm}C{1.2cm}C{1.2cm}C{1.2cm}C{1.2cm}C{1.2cm}C{1.2cm}|C{1.2cm}C{1.2cm}C{1.2cm}}
\hline
\hline
~
&\multicolumn{7}{c|}{Given air-light, no post-process}
&\multicolumn{3}{c}{Configuration of Berman~\cite{berman2016non}}\\
\hline
Metrics
& DC~\cite{he2011single}
& Berman~\cite{berman2016non}
& Meng~\cite{meng2013efficient}
& Ren~\cite{ren2016single}
& Zhu~\cite{Zhu2017Single}
& WDC
& CWDC
& Berman~\cite{berman2016non}
& WDC
& CWDC\\
\hline
MSE      &0.235  &0.242  &0.218  &0.313 &\cellcolor{blue!10}0.215  &\cellcolor{blue!20}0.213  &\cellcolor{blue!30}0.206
&\cellcolor{blue!10}0.284 &\cellcolor{blue!20}0.237 &\cellcolor{blue!30}0.230 \tabularnewline
CIEDE2000&11.190 &11.276 &10.224 &15.064 &\cellcolor{blue!10}10.097 &\cellcolor{blue!20}9.994  &\cellcolor{blue!30}9.687
&\cellcolor{blue!10}15.618&\cellcolor{blue!20}13.416&\cellcolor{blue!30}12.977\tabularnewline
\hline
SSIM     &0.836  &0.806  &0.837  &0.802 &\cellcolor{blue!20}0.847  &\cellcolor{blue!10}0.846  &\cellcolor{blue!30}0.853
&\cellcolor{blue!10}0.775 &\cellcolor{blue!20}0.835 &\cellcolor{blue!30}0.844 \tabularnewline
MS-SSIM  &0.819  &0.834  &0.837  &0.789 &\cellcolor{blue!30}0.846  &\cellcolor{blue!10}0.837  &\cellcolor{blue!20}0.842
&\cellcolor{blue!10}0.798 &\cellcolor{blue!20}0.827 &\cellcolor{blue!30}0.834 \tabularnewline
VIF      &0.568  &0.638  &\cellcolor{blue!30}0.667 &0.557 &\cellcolor{blue!10}0.652  &0.647  &\cellcolor{blue!20}0.653
&\cellcolor{blue!10}0.651 &\cellcolor{blue!20}0.663 &\cellcolor{blue!30}0.670 \tabularnewline
IW-SSIM  &0.781  &0.801  &0.800  &0.739 &\cellcolor{blue!30}0.812  &\cellcolor{blue!10}0.803  &\cellcolor{blue!20}0.808
&\cellcolor{blue!10}0.763 &\cellcolor{blue!20}0.793 &\cellcolor{blue!30}0.801 \tabularnewline
RFSIM    &0.963  &0.966  &\cellcolor{blue!20}0.967 &0.949 &\cellcolor{blue!30}0.968  &0.965  &\cellcolor{blue!20}0.967
&\cellcolor{blue!10}0.949 &\cellcolor{blue!20}0.962 &\cellcolor{blue!30}0.965 \tabularnewline
FSIM     &0.896  &0.886  &0.893  &0.884 &\cellcolor{blue!20}0.902  &\cellcolor{blue!10}0.901  &\cellcolor{blue!30}0.906
&\cellcolor{blue!10}0.869 &\cellcolor{blue!20}0.896 &\cellcolor{blue!30}0.902 \tabularnewline
\hline
\hline
\end{tabular}
\\
~\\
~\\
\centering
\caption{\textbf{Comparison on the improved dataset of Fattal~\cite{fattal2014dehazing}.}}
\label{Table:Fattal}
\begin{tabular}{C{1.2cm}|C{1.2cm}C{1.2cm}C{1.2cm}C{1.2cm}C{1.2cm}C{1.2cm}C{1.2cm}|C{1.2cm}C{1.2cm}C{1.2cm}}
\hline
\hline
~
&\multicolumn{7}{c|}{Given air-light, no post-process}
&\multicolumn{3}{c}{Configuration of Berman~\cite{berman2016non}}\\
\hline
~
& DC~\cite{he2011single}
& Berman~\cite{berman2016non}
& Meng~\cite{meng2013efficient}
& Ren~\cite{ren2016single}
& Zhu~\cite{Zhu2017Single}
& WDC
& CWDC
& Berman~\cite{berman2016non}
& WDC
& CWDC
\\
\hline
MSE      &0.110  &0.154  &0.114 &0.161 &\cellcolor{blue!10}0.108  &\cellcolor{blue!20}0.105 &\cellcolor{blue!30}0.098
&\cellcolor{blue!10}0.160 &\cellcolor{blue!20}0.118 &\cellcolor{blue!30}0.113 \tabularnewline
CIEDE2000&5.529  &7.396  &5.558 &7.979 &\cellcolor{blue!10}5.309  &\cellcolor{blue!20}5.173 &\cellcolor{blue!30}4.862
&\cellcolor{blue!10}9.294 &\cellcolor{blue!20}7.562 &\cellcolor{blue!30}7.309 \tabularnewline
\hline
SSIM     &0.929  &0.888  &0.929 &0.890 &\cellcolor{blue!20}0.941  &\cellcolor{blue!10}0.936 &\cellcolor{blue!30}0.944
&\cellcolor{blue!10}0.885 &\cellcolor{blue!20}0.931 &\cellcolor{blue!30}0.938 \tabularnewline
MS-SSIM  &0.929  &0.919  &0.932 &0.904 &\cellcolor{blue!30}0.941  &\cellcolor{blue!10}0.933 &\cellcolor{blue!20}0.939
&\cellcolor{blue!10}0.911 &\cellcolor{blue!20}0.929 &\cellcolor{blue!30}0.934 \tabularnewline
VIF      &0.690  &0.693  &\cellcolor{blue!20}0.734 &0.649 &0.730  &\cellcolor{blue!10}0.731 &\cellcolor{blue!30}0.739
&\cellcolor{blue!10}0.715 &\cellcolor{blue!20}0.748 &\cellcolor{blue!30}0.756\tabularnewline
IW-SSIM  &0.918  &0.910  &\cellcolor{blue!10}0.922 &0.889 &\cellcolor{blue!30}0.931  &\cellcolor{blue!10}0.922 &\cellcolor{blue!20}0.928
&\cellcolor{blue!10}0.900 &\cellcolor{blue!20}0.917 &\cellcolor{blue!30}0.923\tabularnewline
RFSIM    &0.986  &0.984  &0.987 &0.978 &\cellcolor{blue!30}0.989  &\cellcolor{blue!20}0.988 &\cellcolor{blue!30}0.989
&\cellcolor{blue!10}0.979 &\cellcolor{blue!20}0.986 &\cellcolor{blue!30}0.987\tabularnewline
FSIM     &0.953  &0.934  &0.952 &0.933 &\cellcolor{blue!20}0.958  &\cellcolor{blue!10}0.954 &\cellcolor{blue!30}0.960
&\cellcolor{blue!10}0.927 &\cellcolor{blue!20}0.949 &\cellcolor{blue!30}0.954\tabularnewline
\hline
\hline
\end{tabular}
\\
~\\
~\\
\centering
\caption{\textbf{Comparison with end-to-end methods with default configurations.}}
\label{Table:End-to-end}
\begin{tabular}{C{1.2cm}|C{1.2cm}C{1.2cm}C{1.2cm}C{1.2cm}C{1.2cm}|C{1.2cm}C{1.2cm}C{1.2cm}C{1.2cm}C{1.2cm}}
\hline
\hline
~
&\multicolumn{5}{c|}{D-HAZY dataset~\cite{ancuti2016d}}
&\multicolumn{5}{c}{The improved dataset of Fattal~\cite{fattal2014dehazing}}\\
\hline
~
& AOD~\cite{Li2017AOD}
& Zhang~\cite{zhang2018densely}
& Cai~\cite{cai2016dehazenet}
& WDC
& CWDC
& AOD~\cite{Li2017AOD}
& Zhang~\cite{zhang2018densely}
& Cai~\cite{cai2016dehazenet}
& WDC
& CWDC\\
\hline
MSE      &0.336  &0.358  &\cellcolor{blue!10}0.311  &\cellcolor{blue!30}0.270 &\cellcolor{blue!30}0.270
         &0.194 &0.263 &\cellcolor{blue!10}0.136 &\cellcolor{blue!20}0.121 &\cellcolor{blue!30}0.118 \tabularnewline
CIEDE2000&16.701 &17.481 &\cellcolor{blue!10}14.867 &\cellcolor{blue!20}13.081 &\cellcolor{blue!30}13.035
         &10.402&12.890&\cellcolor{blue!10}6.759 &\cellcolor{blue!20}6.176 &\cellcolor{blue!30}6.054 \tabularnewline
\hline
SSIM     &0.776  &0.757  &\cellcolor{blue!10}0.810  &\cellcolor{blue!20}0.823 &\cellcolor{blue!30}0.829
         &0.839 &0.817 &\cellcolor{blue!10}0.900 &\cellcolor{blue!20}0.928 &\cellcolor{blue!30}0.933 \tabularnewline
MS-SSIM  &0.764  &0.774  &\cellcolor{blue!10}0.796  &\cellcolor{blue!20}0.816  &\cellcolor{blue!30}0.823
         &0.868 &0.857 &\cellcolor{blue!10}0.916 &\cellcolor{blue!20}0.930 &\cellcolor{blue!30}0.935 \tabularnewline
VIF      &0.488  &0.478  &\cellcolor{blue!10}0.553  &\cellcolor{blue!20}0.657  &\cellcolor{blue!30}0.662
         &0.544 &0.593 &\cellcolor{blue!10}0.684 &\cellcolor{blue!20}0.727 &\cellcolor{blue!30}0.734 \tabularnewline
IW-SSIM  &0.709  &0.731  &\cellcolor{blue!10}0.747  &\cellcolor{blue!20}0.781  &\cellcolor{blue!30}0.789
         &0.848 &0.839 &\cellcolor{blue!10}0.901 &\cellcolor{blue!20}0.919 &\cellcolor{blue!30}0.925 \tabularnewline
RFSIM    &0.943  &0.937  &\cellcolor{blue!10}0.955  &\cellcolor{blue!20}0.958  &\cellcolor{blue!30}0.961
         &0.968 &0.957 &\cellcolor{blue!10}0.980 &\cellcolor{blue!20}0.988 &\cellcolor{blue!30}0.989 \tabularnewline
FSIM     &0.855  &0.866  &\cellcolor{blue!10}0.887  &\cellcolor{blue!20}0.893  &\cellcolor{blue!30}0.899
         &0.898 &0.887 &\cellcolor{blue!10}0.946 &\cellcolor{blue!20}0.947 &\cellcolor{blue!30}0.952 \tabularnewline
\hline
\hline
\end{tabular}
\\
~\\
~\\
\centering
\caption{\textbf{Mean runtime of each method in quantitative comparison.}}
\label{Table:Runtime}
\begin{tabular}{C{1.2cm}C{1.2cm}C{1.2cm}C{1.2cm}C{1.2cm}C{1.2cm}C{1.2cm}C{1.2cm}C{1.2cm}C{1.2cm}C{1.2cm}}
\hline
\hline
~
& DC~\cite{he2011single}
& Berman~\cite{berman2016non}
& Meng~\cite{meng2013efficient}
& Ren~\cite{ren2016single}
& AOD~\cite{Li2017AOD}
& Zhang~\cite{zhang2018densely}
& Cai~\cite{cai2016dehazenet}
& Zhu~\cite{Zhu2017Single}
& WDC
& CWDC\\
\hline
Platform  &Matlab &Matlab &Matlab &Matlab &PyCaffe  &Pytorch  &Matlab\&C &Matlab &Matlab &Matlab \tabularnewline
withGPU       &-      &-      &-      &-      &$\surd$ &$\surd$ &-          &-      &-      &-      \tabularnewline
Seconds   &8.53   &1.86   &1.27   &1.22   &0.02     &0.02     &2.13       &28.48  &1.47   &21.16  \tabularnewline
\hline
\hline
\end{tabular}
\end{table*}

Some researches mistake the effect of other technologies, such as post-enhancement or
deblurring, as the effect of haze removal part.
As the example displayed in Fig.~\ref{fig:Post},
along with the benefit of post-enhancement, any result on the right column could surpass all the results in the left column in the view of visual pleasuring.
Such comparison is unfair and ineffective.
In most of our comparisons, we exclude post-enhancement and uniform air-light estimates to avoid a targeted downgrading or promoting,
and focus on the differences on transmission estimation.

Several outstanding methods with available codes are selected, including
He~\emph{et~al.}~\cite{he2011single}, Berman~\emph{et~al.}~\cite{berman2016non}, Meng~\emph{et~al.}~\cite{meng2013efficient} and Zhu~\emph{et~al.}~\cite{Zhu2017Single}.
Learning-based methods include Ren~\emph{et~al.}~\cite{ren2016single}, Li~\emph{et~al.}~\cite{Li2017AOD}, Zhang and Patel~\cite{zhang2018densely}, and Cai~\cite{cai2016dehazenet}.
Inputs are scaled so that the maximum of width or height is 640 pixels.
For dark channel prior based methods~\cite{he2011single,meng2013efficient,Zhu2017Single}, a round mask $\Omega$ is used, and its radius is set as 25 pixels.
For all the methods including transmission compensation, Eq.~(\ref{Eq:dehaze}) is used and $\epsilon_t=0.05$.

\subsection{Quantitative results}
MSE (mean square error), CIEDE2000~\cite{Sharma2010The}, SSIM~\cite{wang2004image} and MS-SSIM~\cite{wang2003multiscale} are employed as metrics.
Moreover, we use several outstanding metrics in Zhang~\emph{et~al.}~\cite{zhang2012comprehensive}, including
VIF~\cite{sheikh2006image}, IW-SSIM~\cite{wang2011information}, RFSIM~\cite{zhang2010rfsim} and FSIM~\cite{zhang2011fsim}.
The datasets include D-Hazy~\cite{ancuti2016d} and
the one in Zhu~\emph{et~al.}~\cite{Zhu2017Single},
which is an improved version of the dataset in Fattal~\cite{fattal2014dehazing}.

In the first comparison, ground-truth air-light estimates are provided to each method.
The results are displayed in the left parts of Table.~\ref{Table:D-HAZY} and Table.~\ref{Table:Fattal}\footnote{
Note that, $\epsilon_t$ equals $0.1$ in our previous work~\cite{Zhu2017Single},
which reports globally higher scores.
However, we found out that $\epsilon_t=0.1$ is slightly abused, introducing obvious haze preservation in many cases.
In this paper, we use a more reasonable value, which is also closer to the ones in He~\emph{et~al.}~\cite{he2011single} (0.05 in the paper) and Berman~\emph{et~al.}~\cite{berman2016non} (0.06 in the code).
We also updated our previous experiment.
Please check our project page for the performances of WDC and CWDC under $\epsilon_t=0.1$.
The conclusion is the same.
{\url{https://jiantaoliu.github.io/WDC/}}
}.
It is obvious that the top 3 methods are CWDC, Zhu~\emph{et~al.}~\cite{Zhu2017Single} (our previous work) and WDC.
On the local maximums of lower bound, WDC and CWDC have high weights,
similarly, Zhu~\emph{et~al.}~\cite{Zhu2017Single} have strong data constraint because the label sets there contain few elements.
They are all providing the most smooth transmission map under the constraints of dark pixels.
An interesting fact is that the accuracy of their solutions are also ranked in this way.
WDC neglects the lower bound constraint.
Zhu~\emph{et~al.}~\cite{Zhu2017Single} considers the constraint, but the precision is degraded by color bit down-sampling.
CWDC follows the constraint precisely.

Providing air-light and disabling post-enhancement might be unfair to Berman~\emph{et~al.}~\cite{berman2016non}, because
air-light estimation algorithm based on haze-line~\cite{berman2017air} has been reported as benefit to its performance,
and contrast-stretch might be specially suitable for its brightness imbalance.
Berman~\emph{et~al.}~\cite{berman2016non} might has better rank under its default configuration.
Therefore, in our second comparison, WDC and CWDC follow the default configuration of the code provided by Berman~\emph{et~al.}~\cite{berman2016non},
including haze-line based air-light estimation and contrast-stretch based post-enhancement.
The results are displayed in right parts of Table.~\ref{Table:D-HAZY} and Table.~\ref{Table:Fattal}.
As shown, our methods still have a superior performance.

Li~\emph{et~al.}~\cite{Li2017AOD}, Zhang and Patel~\cite{zhang2018densely},
\footnote{
Zhang and Patel~\cite{zhang2018densely} only receives and produces 512x512 images.
As the authors suggested, the inputs are resized as 512x512 and output are resized as 640x480.
It might be unfair since the results are up-sampled in vertical direction.
Please check our project page for the comparison on 512x512 inputs and outputs.
The conclusion is the same.
{\url{https://jiantaoliu.github.io/WDC/}}}
and Cai~\emph{et~al.}~\cite{cai2016dehazenet} are end-to-end.
Therefore, they are compared in another experiment, where all the results are provided under default configurations.
The results are displayed in Table.~\ref{Table:End-to-end}, where WDC and CWDC have obvious superiority.

\begin{figure*}[tp]
\begin{subfigure}{0.19\linewidth}\includegraphics[width=\textwidth,height=0.6\textwidth]{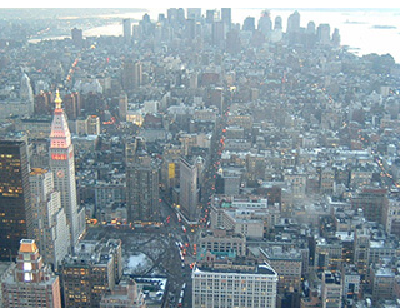}
\end{subfigure}
\begin{subfigure}{0.19\linewidth}\includegraphics[width=\textwidth,height=0.6\textwidth]{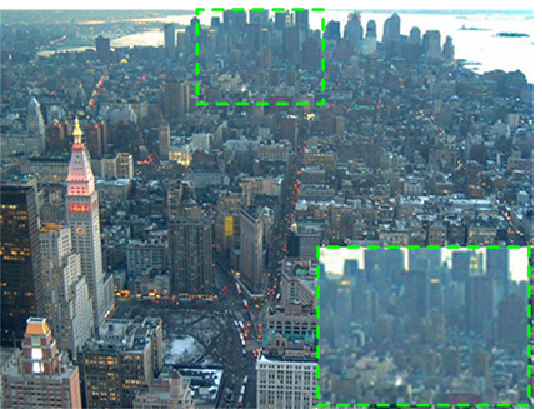}
\end{subfigure}
\begin{subfigure}{0.19\linewidth}\includegraphics[width=\textwidth,height=0.6\textwidth]{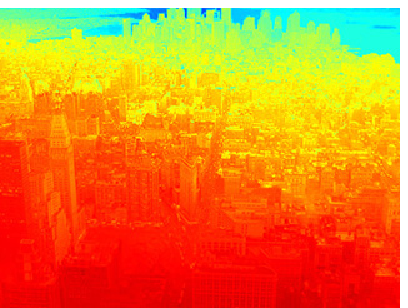}
\end{subfigure}
\begin{subfigure}{0.19\linewidth}\includegraphics[width=\textwidth,height=0.6\textwidth]{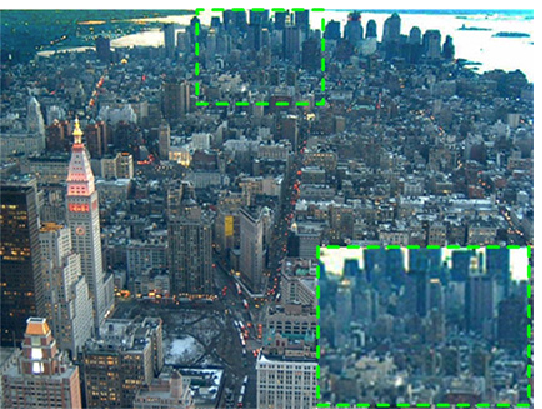}
\end{subfigure}
\begin{subfigure}{0.19\linewidth}\includegraphics[width=\textwidth,height=0.6\textwidth]{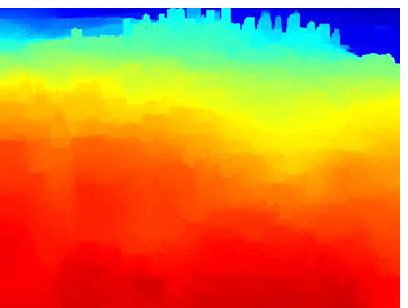}
\end{subfigure}
\\
\begin{subfigure}{0.19\linewidth}\includegraphics[width=\textwidth,height=0.6\textwidth]{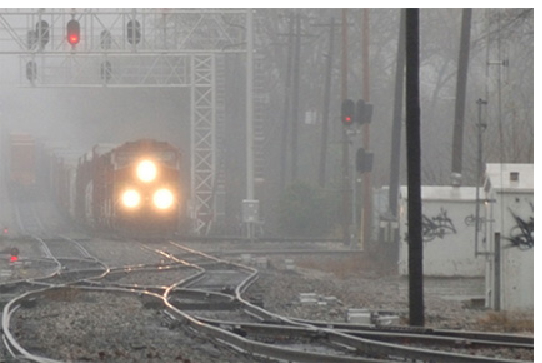}
\caption{input images}
\end{subfigure}
\begin{subfigure}{0.19\linewidth}\includegraphics[width=\textwidth,height=0.6\textwidth]{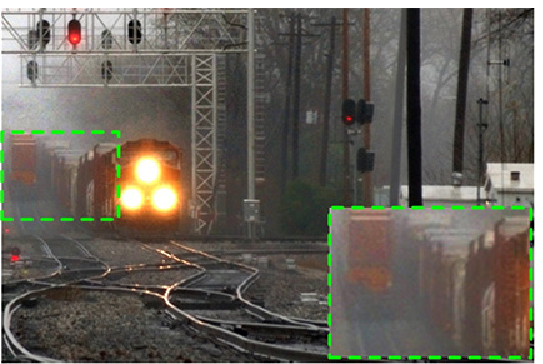}
\caption{He~\emph{et~al.}~\cite{he2011single}}
\end{subfigure}
\begin{subfigure}{0.19\linewidth}\includegraphics[width=\textwidth,height=0.6\textwidth]{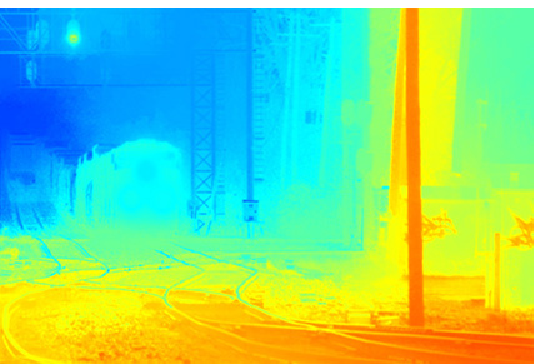}
\caption{trans. map of~\cite{he2011single}}
\end{subfigure}
\begin{subfigure}{0.19\linewidth}\includegraphics[width=\textwidth,height=0.6\textwidth]{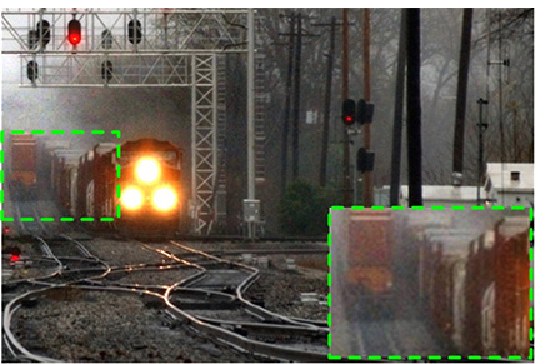}
\caption{WDC}
\end{subfigure}
\begin{subfigure}{0.19\linewidth}\includegraphics[width=\textwidth,height=0.6\textwidth]{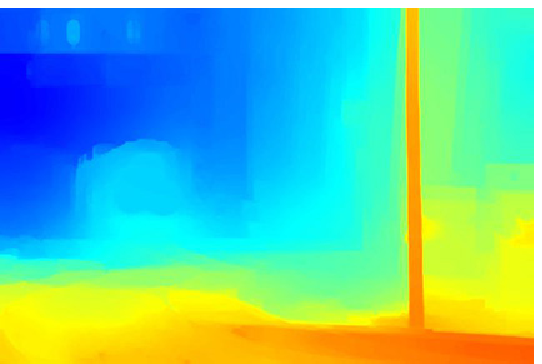}
\caption{trans. map of WDC}
\end{subfigure}
\\
\caption{\textbf{Comparison of WDC and He~\emph{et~al.}~\cite{he2011single}.}}
\label{fig:QChe}
\end{figure*}

\begin{figure*}[tp]
\begin{subfigure}{0.19\linewidth}\includegraphics[width=\textwidth,height=0.6\textwidth]{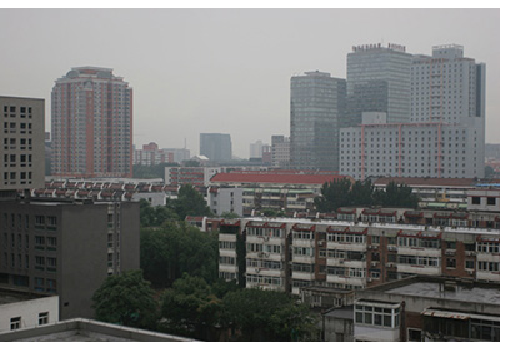}
\end{subfigure}
\begin{subfigure}{0.19\linewidth}\includegraphics[width=\textwidth,height=0.6\textwidth]{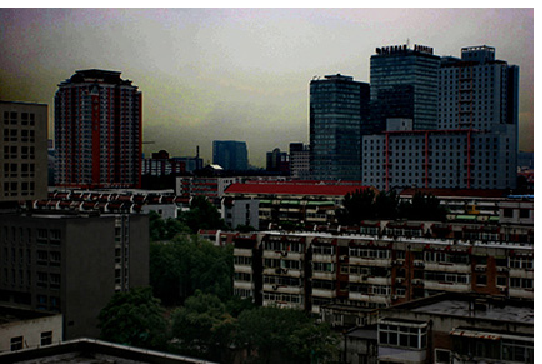}
\end{subfigure}
\begin{subfigure}{0.19\linewidth}\includegraphics[width=\textwidth,height=0.6\textwidth]{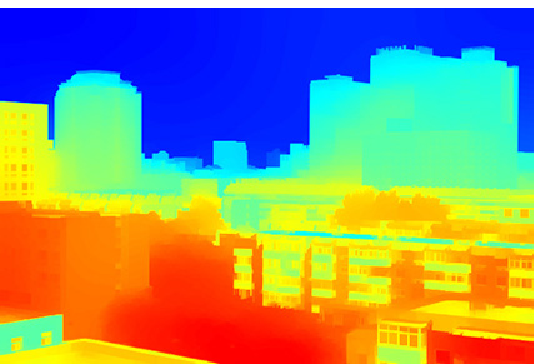}
\end{subfigure}
\begin{subfigure}{0.19\linewidth}\includegraphics[width=\textwidth,height=0.6\textwidth]{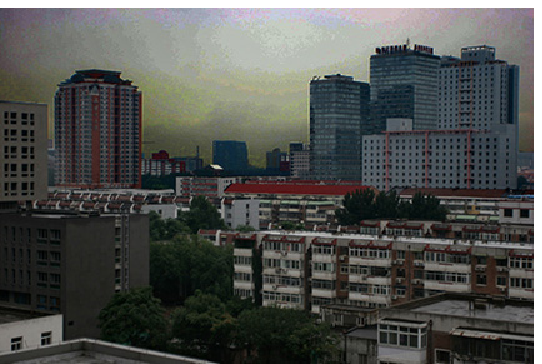}
\end{subfigure}
\begin{subfigure}{0.19\linewidth}\includegraphics[width=\textwidth,height=0.6\textwidth]{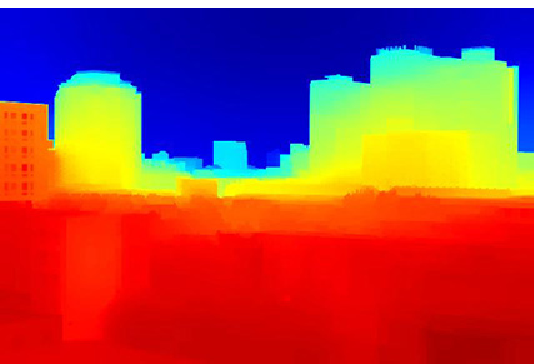}
\end{subfigure}
\\
\begin{subfigure}{0.19\linewidth}\includegraphics[width=\textwidth,height=0.6\textwidth]{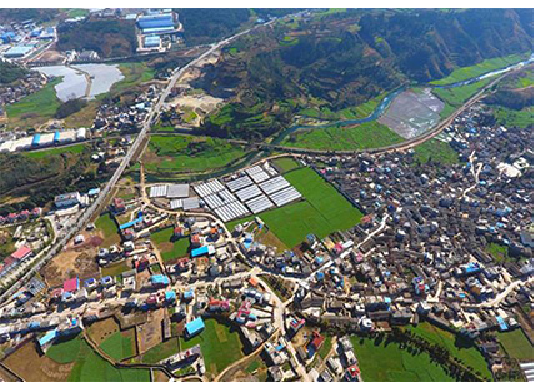}
\caption{input images}
\end{subfigure}
\begin{subfigure}{0.19\linewidth}\includegraphics[width=\textwidth,height=0.6\textwidth]{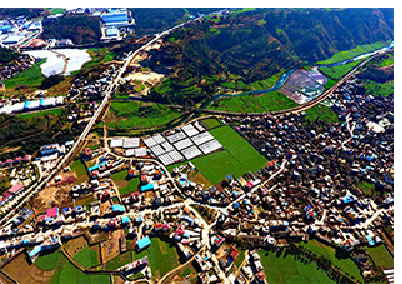}
\caption{Berman~\emph{et~al.}~\cite{berman2016non}}
\label{fig:QCberman-b}
\end{subfigure}
\begin{subfigure}{0.19\linewidth}\includegraphics[width=\textwidth,height=0.6\textwidth]{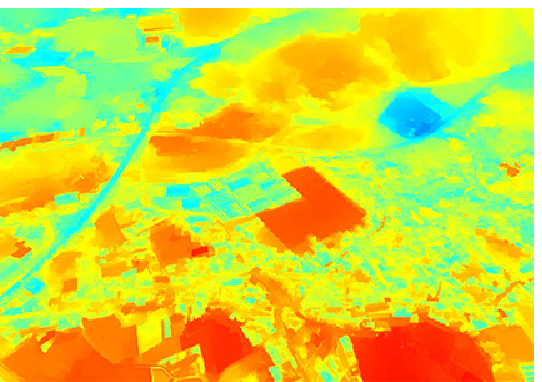}
\caption{trans. map of~\cite{berman2016non}}
\label{fig:QCberman-c}
\end{subfigure}
\begin{subfigure}{0.19\linewidth}\includegraphics[width=\textwidth,height=0.6\textwidth]{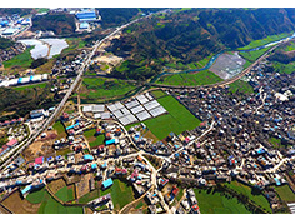}
\caption{WDC}
\end{subfigure}
\begin{subfigure}{0.19\linewidth}\includegraphics[width=\textwidth,height=0.6\textwidth]{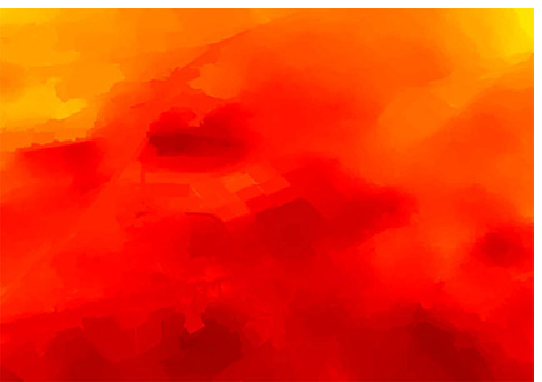}
\caption{trans. map of WDC}
\end{subfigure}
\\
\caption{\textbf{Comparison of WDC and Berman~\emph{et~al.}~\cite{berman2016non}.}}
\label{fig:QCberman}
\end{figure*}

\begin{figure*}[tp]
\begin{subfigure}{0.19\linewidth}\includegraphics[width=\textwidth,height=0.6\textwidth]{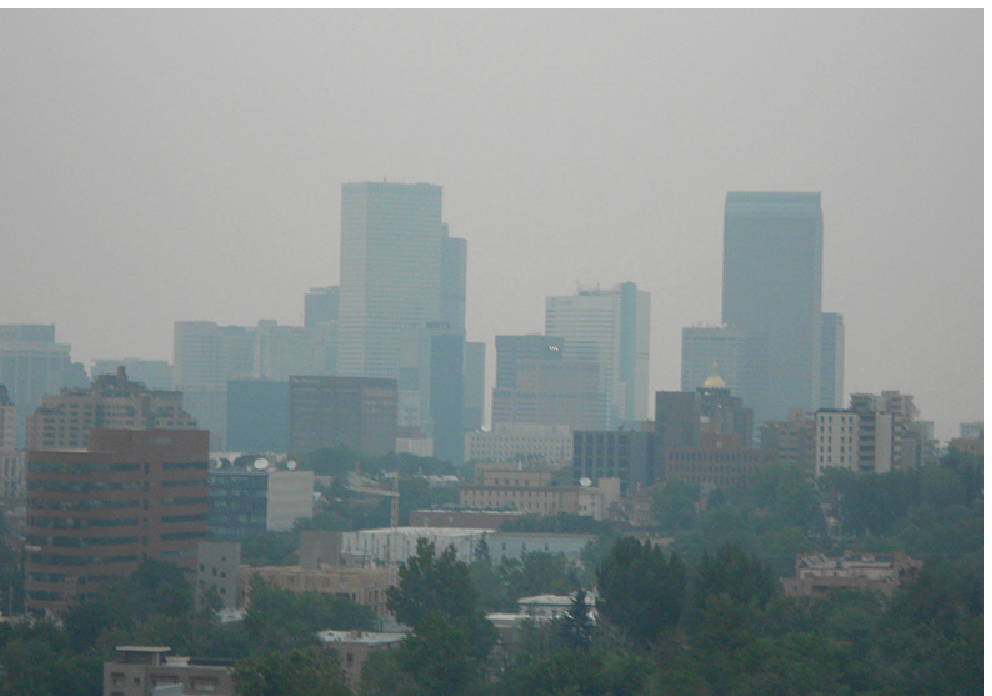}
\end{subfigure}
\begin{subfigure}{0.19\linewidth}\includegraphics[width=\textwidth,height=0.6\textwidth]{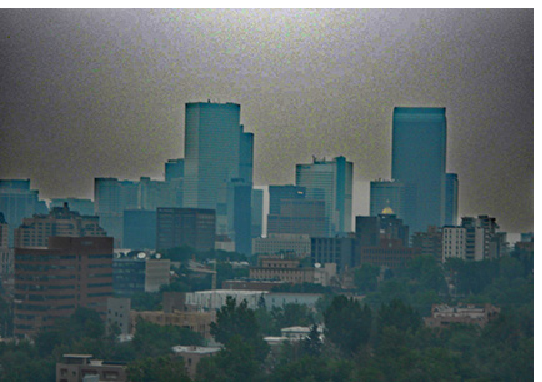}
\end{subfigure}
\begin{subfigure}{0.19\linewidth}\includegraphics[width=\textwidth,height=0.6\textwidth]{QCmeng_meng_T}
\end{subfigure}
\begin{subfigure}{0.19\linewidth}\includegraphics[width=\textwidth,height=0.6\textwidth]{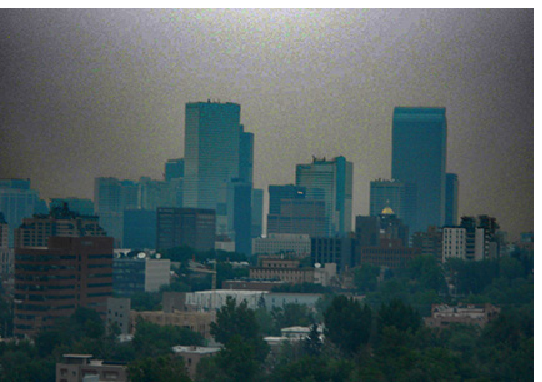}
\end{subfigure}
\begin{subfigure}{0.19\linewidth}\includegraphics[width=\textwidth,height=0.6\textwidth]{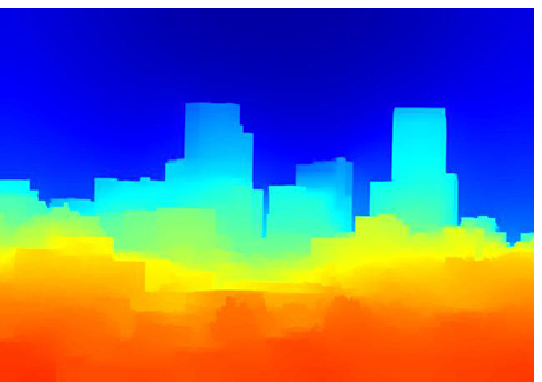}
\end{subfigure}
\\
\begin{subfigure}{0.19\linewidth}\includegraphics[width=\textwidth,height=0.6\textwidth]{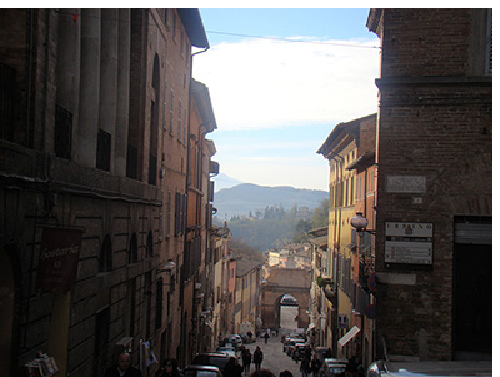}
\caption{input images}
\end{subfigure}
\begin{subfigure}{0.19\linewidth}\includegraphics[width=\textwidth,height=0.6\textwidth]{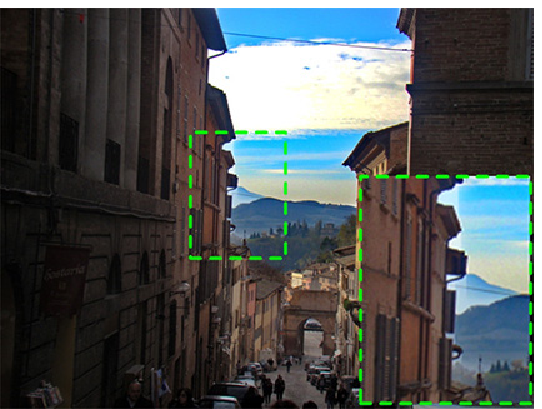}
\caption{Meng~\emph{et~al.}~\cite{meng2013efficient}}
\end{subfigure}
\begin{subfigure}{0.19\linewidth}\includegraphics[width=\textwidth,height=0.6\textwidth]{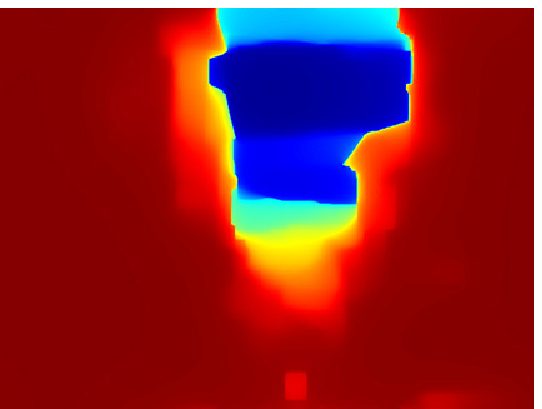}
\caption{trans. map of~\cite{meng2013efficient}}
\end{subfigure}
\begin{subfigure}{0.19\linewidth}\includegraphics[width=\textwidth,height=0.6\textwidth]{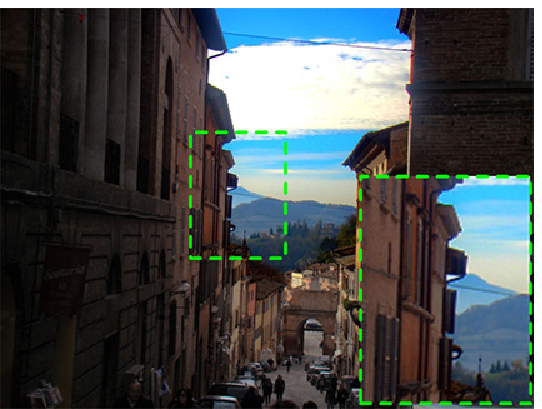}
\caption{WDC}
\end{subfigure}
\begin{subfigure}{0.19\linewidth}\includegraphics[width=\textwidth,height=0.6\textwidth]{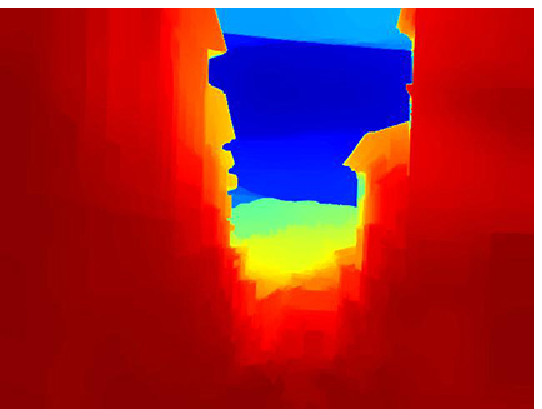}
\caption{trans. map of WDC}
\end{subfigure}
\\
\caption{\textbf{Comparison of WDC and Meng~\emph{et~al.}~\cite{meng2013efficient}.}}
\label{fig:QCmeng}
\end{figure*}

\begin{figure*}[tp]
\begin{subfigure}{0.19\linewidth}\includegraphics[width=\textwidth,height=0.6\textwidth]{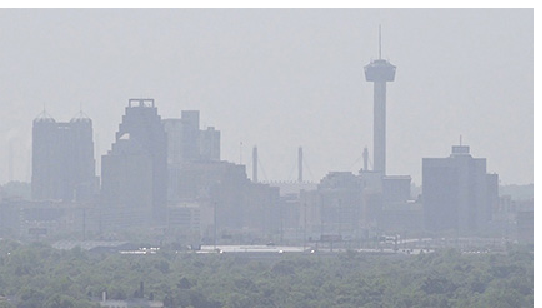}
\end{subfigure}
\begin{subfigure}{0.19\linewidth}\includegraphics[width=\textwidth,height=0.6\textwidth]{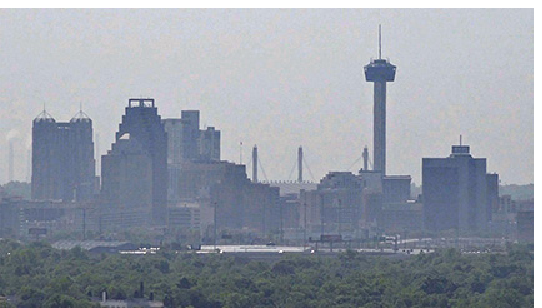}
\end{subfigure}
\begin{subfigure}{0.19\linewidth}\includegraphics[width=\textwidth,height=0.6\textwidth]{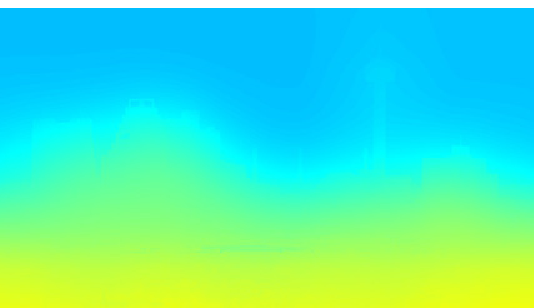}
\end{subfigure}
\begin{subfigure}{0.19\linewidth}\includegraphics[width=\textwidth,height=0.6\textwidth]{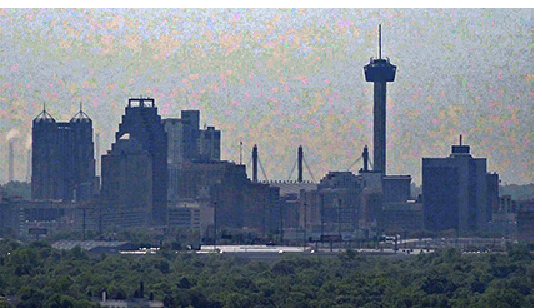}
\end{subfigure}
\begin{subfigure}{0.19\linewidth}\includegraphics[width=\textwidth,height=0.6\textwidth]{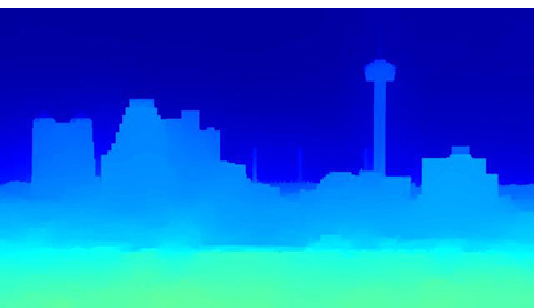}
\end{subfigure}
\\
\begin{subfigure}{0.19\linewidth}\includegraphics[width=\textwidth,height=0.6\textwidth]{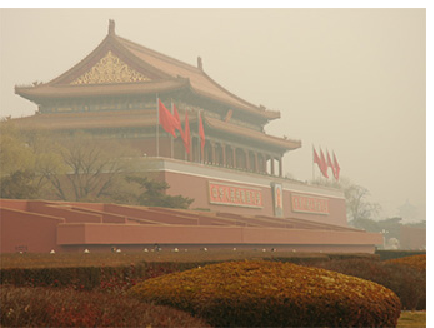}
\caption{input images}
\end{subfigure}
\begin{subfigure}{0.19\linewidth}\includegraphics[width=\textwidth,height=0.6\textwidth]{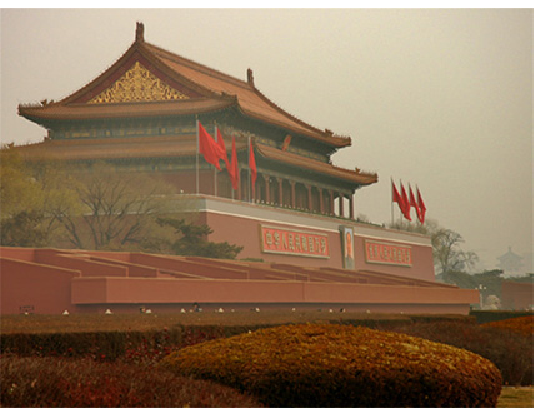}
\caption{Ren~\emph{et~al.}~\cite{ren2016single}}
\end{subfigure}
\begin{subfigure}{0.19\linewidth}\includegraphics[width=\textwidth,height=0.6\textwidth]{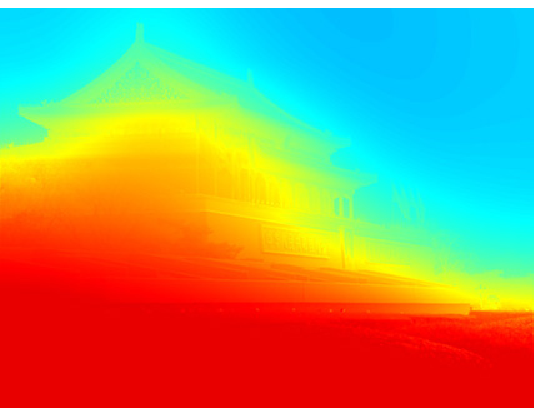}
\caption{trans. map of~\cite{ren2016single}}
\label{subfig:QCren-c}
\end{subfigure}
\begin{subfigure}{0.19\linewidth}\includegraphics[width=\textwidth,height=0.6\textwidth]{QCren2_wdc_J}
\caption{WDC}
\end{subfigure}
\begin{subfigure}{0.19\linewidth}\includegraphics[width=\textwidth,height=0.6\textwidth]{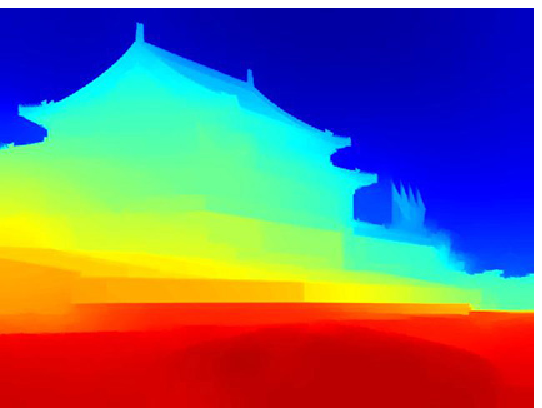}
\caption{trans. map of WDC}
\end{subfigure}
\\
\caption{\textbf{Comparison of WDC and Ren~\emph{et~al.}~\cite{ren2016single}.}}
\label{fig:QCren}
\end{figure*}

\begin{figure*}[t]
\centering
\includegraphics[width=0.83\textwidth]{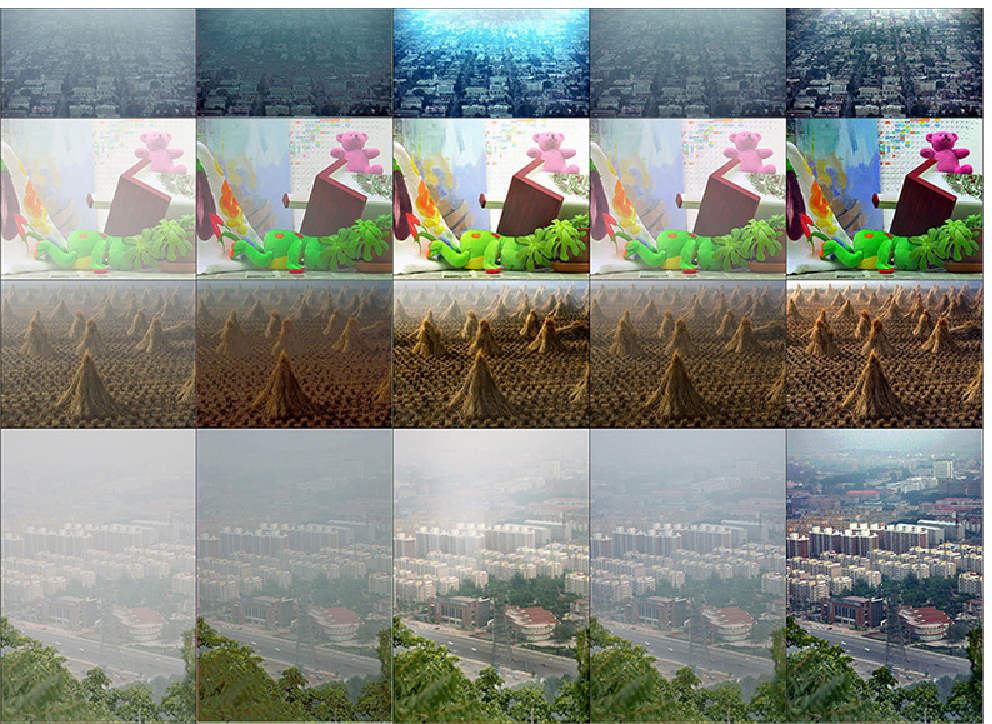}
\\
\caption{\textbf{Comparison of several learning-based end-to-end methods and WDC.}
From left to right, input images, Li~\emph{et~al.}~\cite{Li2017AOD}, Zhang and Patel~\cite{zhang2018densely}, Cai~\emph{et~al.}~\cite{cai2016dehazenet}, and WDC.}
\label{fig:QCe2e}
\end{figure*}

The runtimes of the methods are summarized in Table~\ref{Table:Runtime}.
Learning-based methods require a relatively time-consuming initialization to load the net parameters.
Such costs are excluded in the comparison.
As shown, CWDC is quite slow, while WDC is competitive and even comparable to learning-based methods on only CPU.
Refer to Alg.~\ref{Alg:dehaze}, most processes of WDC are basic operations.
The only time-consuming step is Eq.~(\ref{Eq:refinement-matrix-solution}), which is also fast in a modest image size.
Among non-learning based methods, WDC is inferior to Meng~\emph{et~al.}~\cite{meng2013efficient}.
This gap increases with image size because Meng~\emph{et~al.}~\cite{meng2013efficient} is pixel-wise but WDC requires a sparse matrix inversion.
However, the speed of Meng~\emph{et~al.}~\cite{meng2013efficient} is in the price of missing global optimum.

In summary, the quantitative results demonstrate the superiority of our methods.
Whether compared with learning or non-learning based methods, traditional or end-to-end methods, with or without air-light groundtruth and post-enhancement,
WDC achieves satisfactory results in both quality and speed, and CWDC is top ranked in quality.

\subsection{Qualitative results}

In the first comparison, we compare WDC with the methods containing transmission estimation process.
Air-lights are provided by He~\emph{et~al.}~\cite{he2011single}, and post-enhancements are disabled.
Note that, because of the absent of post-enhancement, some results are darker in this paper than in their original reports,
thus less satisfactory in the view of visual pleasuring.
As an easier evaluation method, which is also more reasonable, is by checking failures in transmission maps and then identifying the consequences(transmissions should well and only reflect scene depth).

Fig.~\ref{fig:QChe} compares WDC with He~\emph{et~al.}~\cite{he2011single}.
The transmission maps are similar in general because of the same prior.
The problem of He~\emph{et~al.}~\cite{he2011single} is the wide existence of depth-irrelevant details in transmission map,
leading to micro-contrast loss, which is evident in the zoomed regions.
As a comparison, WDC well reflects scene depth in the transmission maps and preserves more details.

Fig.~\ref{fig:QCberman} compares WDC with Berman~\emph{et~al.}~\cite{berman2016non}.
The haze-line model of Berman~\emph{et~al.}~\cite{berman2016non} has a major defect.
That is, the colors of haze and white objects are not separated.
Although this ambiguity is recognized as inevitable,
haze-line model might lead to a more serious consequence than the dark channel prior because
it affects the result globally.
As shown in Fig.~\ref{fig:QCberman-c}, the model fails completely due to the white objects being
recognized as haze, resulting in the chaotic transmission maps.
As a comparison, WDC performs stably.

Fig.~\ref{fig:QCmeng} compares WDC with Meng~\emph{et~al.}~\cite{meng2013efficient}.
The transmission maps of Meng~\emph{et~al.}~\cite{meng2013efficient} are over-smoothed
because of the reason we discussed in Section~\ref{sec:introduction}.
On zigzag depth discontinuities, the over-smoothed transmission estimates result in halo-effect.
As a comparison, WDC well reflects these discontinuities and is free from the halo-effect.

Fig.~\ref{fig:QCren} compares WDC with Ren~\emph{et~al.}~\cite{ren2016single}, which estimates transmissions by training.
It appears that Ren~\emph{et~al.}~\cite{ren2016single} estimates transmissions in a rough way.
The haze distributions are generally reflected but depth discontinuities are all degraded.
Furthermore, transmissions are over-estimated thus haze are not completely removed.

In the second comparison, WDC is compared with end-to-end methods including Li~\emph{et~al.}~\cite{Li2017AOD}, Zhang and Patel~\cite{zhang2018densely}, and Cai~\emph{et~al.}~\cite{cai2016dehazenet}.
The results are all provided under the default configurations.
Fig.~\ref{fig:QCe2e} shows the results,
where problems of haze preservation, micro-contrast loss, color bias exist in the three methods while
WDC produces relatively satisfactory results
(more examples can be found in the projectpage, \url{https://jiantaoliu.github.io/WDC/}).

\section{EWDC}
\label{sec:EWDC}

\begin{figure}[t]
\centering
\begin{subfigure}{0.9\linewidth}
\includegraphics[width=0.3\textwidth,height=0.3\textwidth]{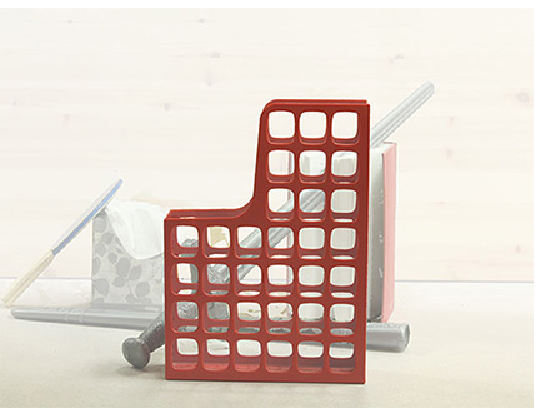}
\includegraphics[width=0.3\textwidth,height=0.3\textwidth]{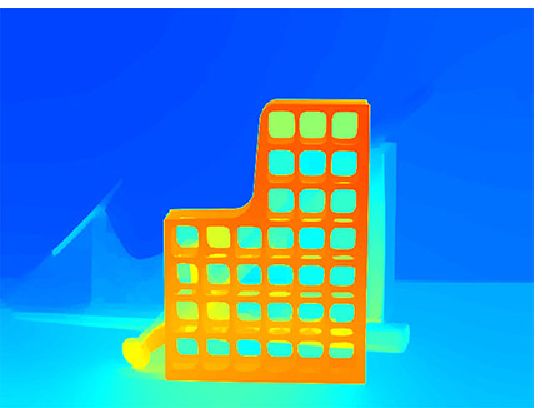}
\includegraphics[width=0.3\textwidth,height=0.3\textwidth]{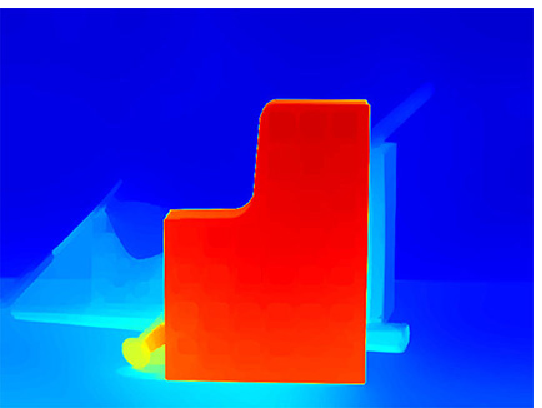}
\end{subfigure}
\centering
\begin{subfigure}{0.9\linewidth}
\includegraphics[width=0.3\textwidth,height=0.3\textwidth]{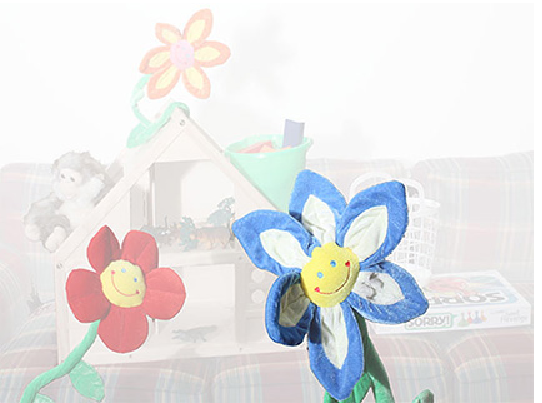}
\includegraphics[width=0.3\textwidth,height=0.3\textwidth]{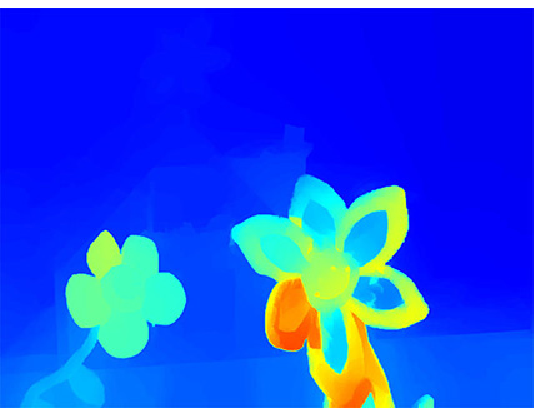}
\includegraphics[width=0.3\textwidth,height=0.3\textwidth]{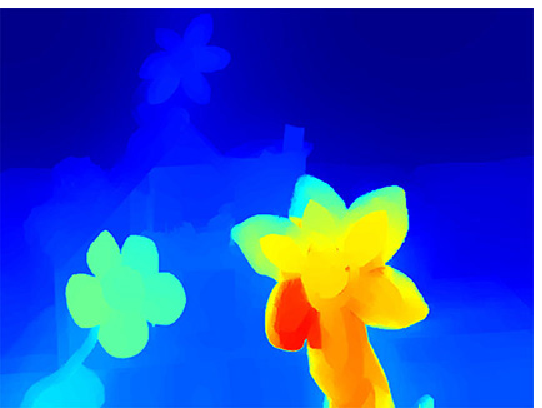}
\end{subfigure}
\\
\caption{\textbf{Comparison of failure cases of Berman~\emph{et~al.}~\cite{berman2016non} and WDC (middle and bottom rows repectively).}}
\label{fig:Ambiguity-Example}
\end{figure}

\begin{figure}[t]
\begin{subfigure}{0.30\linewidth}\includegraphics[width=\textwidth,height=0.6\textwidth]{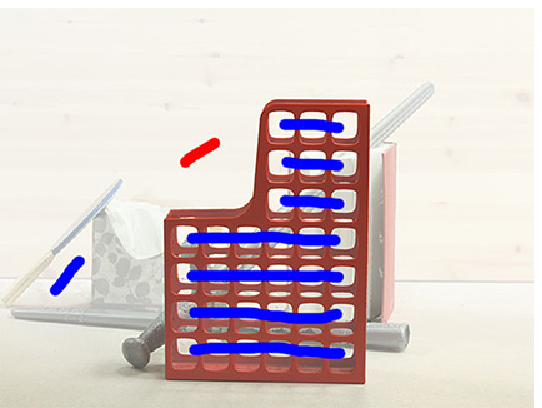}
\end{subfigure}
\begin{subfigure}{0.30\linewidth}\includegraphics[width=\textwidth,height=0.6\textwidth]{Amb_wdc}
\end{subfigure}
\begin{subfigure}{0.3\linewidth}\includegraphics[width=\textwidth,height=0.6\textwidth]{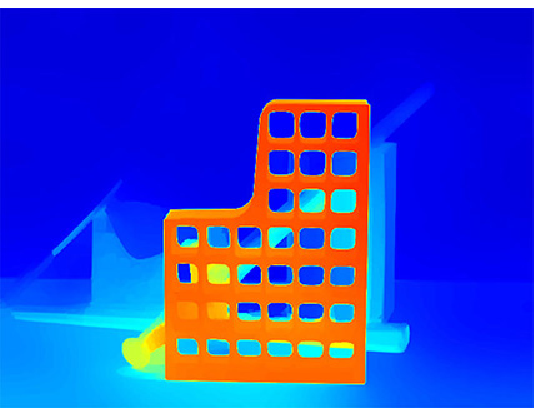}
\end{subfigure}
\\
\begin{subfigure}{0.3\linewidth}\includegraphics[width=\textwidth,height=0.6\textwidth]{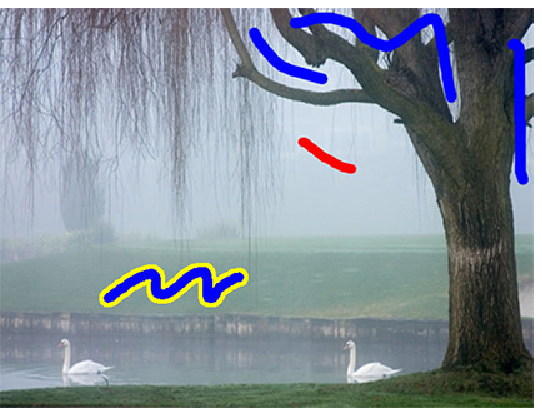}
\end{subfigure}
\begin{subfigure}{0.3\linewidth}\includegraphics[width=\textwidth,height=0.6\textwidth]{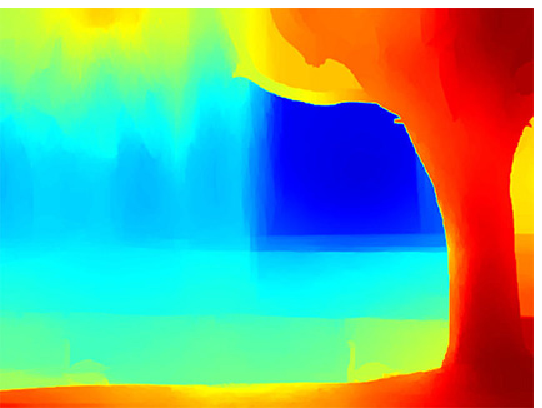}
\end{subfigure}
\begin{subfigure}{0.3\linewidth}\includegraphics[width=\textwidth,height=0.6\textwidth]{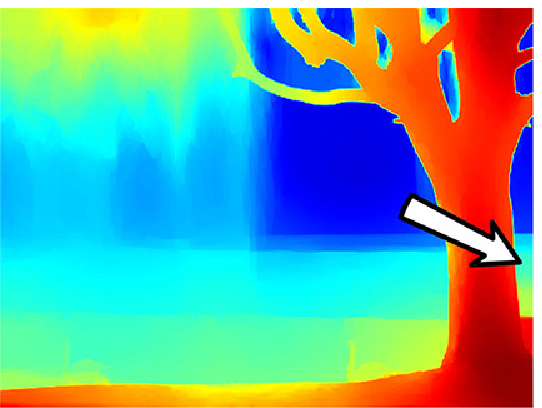}
\end{subfigure}
\\
\begin{subfigure}{0.3\linewidth}\includegraphics[width=\textwidth,height=0.6\textwidth]{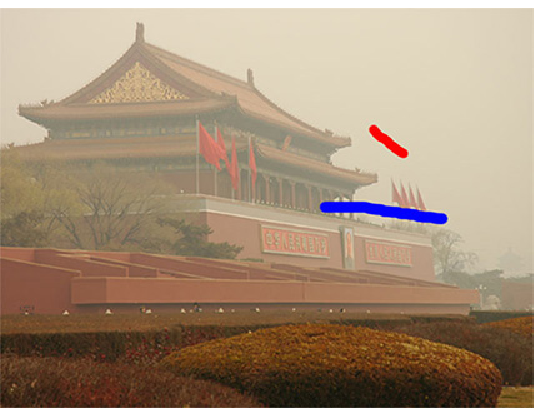}
\end{subfigure}
\begin{subfigure}{0.3\linewidth}\includegraphics[width=\textwidth,height=0.6\textwidth]{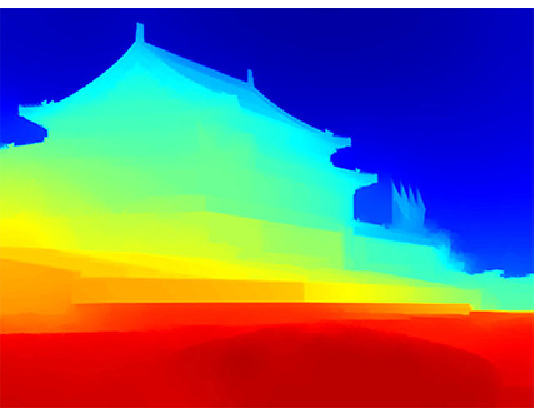}
\end{subfigure}
\begin{subfigure}{0.3\linewidth}\includegraphics[width=\textwidth,height=0.6\textwidth]{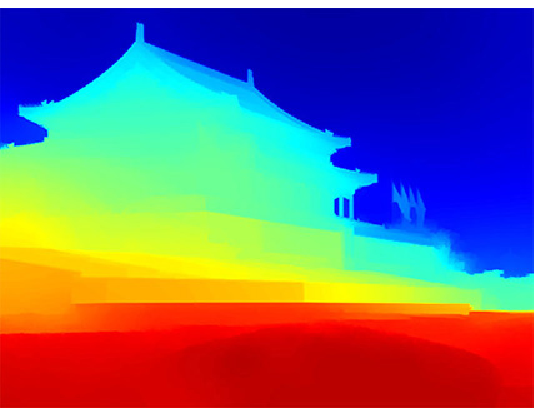}
\end{subfigure}
\\
\begin{subfigure}{0.3\linewidth}\includegraphics[width=\textwidth,height=0.6\textwidth]{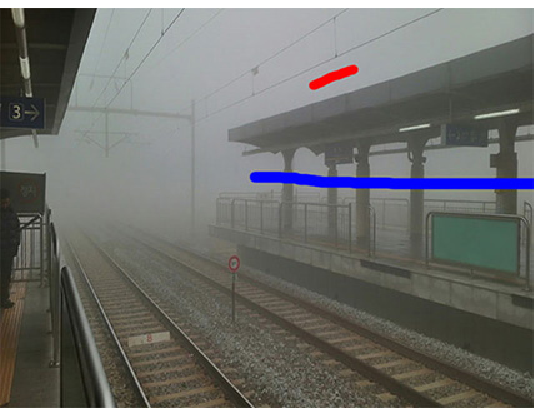}
\caption{inputs}
\end{subfigure}
\begin{subfigure}{0.3\linewidth}\includegraphics[width=\textwidth,height=0.6\textwidth]{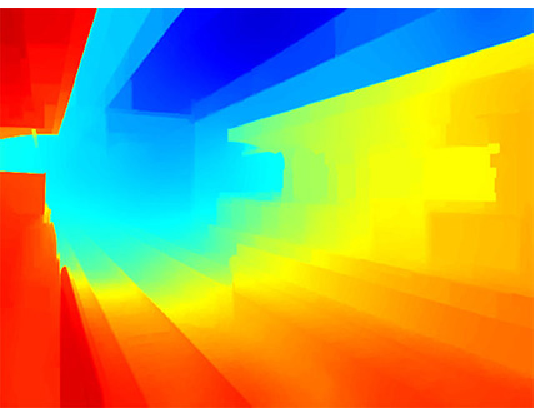}
\caption{WDC}
\end{subfigure}
\begin{subfigure}{0.3\linewidth}\includegraphics[width=\textwidth,height=0.6\textwidth]{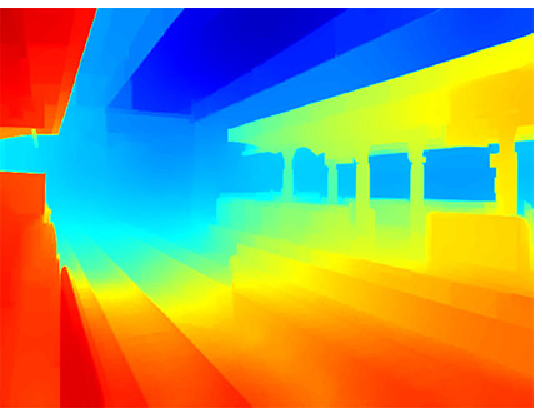}
\caption{EWDC}
\end{subfigure}
\\
\caption{\textbf{An instance of EWDC with hand-drawn messages.}}
\label{fig:Com-RefineSample}
\end{figure}

\begin{figure}[t]
\centering
\begin{subfigure}{0.4\linewidth}\includegraphics[width=\textwidth,height=0.6\textwidth]{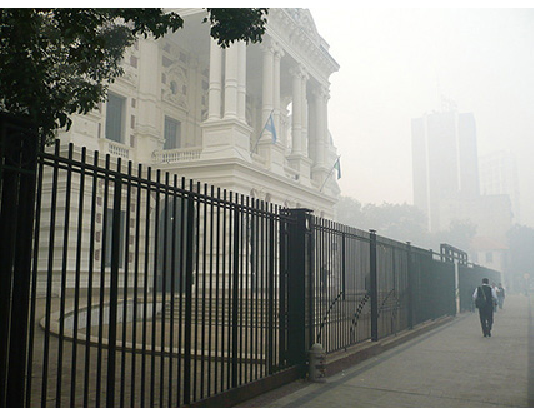}
\caption{input}
\end{subfigure}
\begin{subfigure}{0.4\linewidth}\includegraphics[width=\textwidth,height=0.6\textwidth]{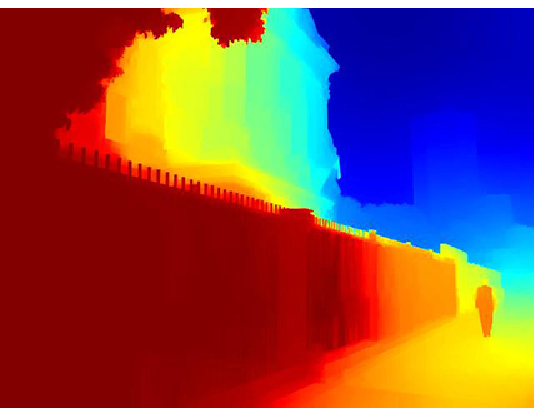}
\caption{WDC}
\end{subfigure}
\\
\centering
\begin{subfigure}{0.4\linewidth}\includegraphics[width=\textwidth,height=0.6\textwidth]{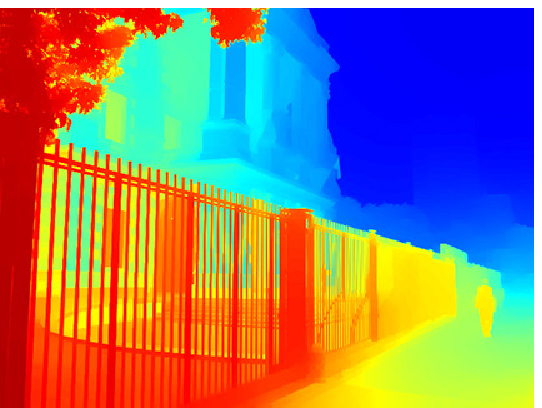}
\caption{Berman~\emph{et~al.}~\cite{berman2016non}}
\end{subfigure}
\begin{subfigure}{0.4\linewidth}\includegraphics[width=\textwidth,height=0.6\textwidth]{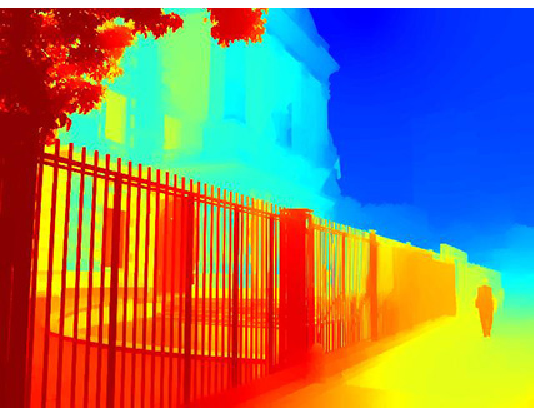}
\caption{EWDC}
\end{subfigure}
\\
\caption{\textbf{An instance of EWDC with a global assumption.}}
\label{fig:EWDC-assume}
\end{figure}

EWDC (extensible WDC) asks for extra messages to rediscover the missing dark pixels which are essential to the transmission estimation but neglected by WDC due to the limitation of preset masks.
For example, the hollows of the shelf in Fig.~\ref{fig:Ambiguity-Example} are covered by the masks $\Omega$ thus no dark pixel is detected.
As a consequence, the transmissions of these isolated backgrounds are over-estimated as the values of the shelf.
Berman~\emph{et~al.}~\cite{berman2016non} performs better in this sample since these isolated backgrounds and the major one are classified together by its haze-line model.
WDC can also produce the right estimates by reducing the size of $\Omega$.
However, small $\Omega$ will undermine the dark channel prior, recognizing the white part of the flower as background and resulting in the same under-estimated failure likes Berman~\emph{et~al.}~\cite{berman2016non}.
Apparently, there is no universally correct $\Omega$.
External knowledge is required for such cases.

We define an interface to receive external knowledge, which should be transformed into messages defined as a structure of set $C$ and value $t_s$.
The set $C$ restores the pixels that are supposed to have the same transmission $t_s$, which can be $\max_{y\in{C}}b(y)$ or provided specifically.
With each message, we modify the initial transmission map as $\tilde{t}(y)=t_s$ for $y\in{C}$, and recalculate the weight map and the final estimates $t$.
This process seems aggressive and requires the extra messages being precise, but it is actually robust as long as key dark pixels are located.
We provide two instances as examples of EWDC.

The first instance is the application illustrated in Fig.~\ref{fig:Com-RefineSample},
where extra messages are provided by scribbles.
The blue marks indicate $C$ and the red mark picks $t_s$.
As shown, the blue marks only need to strike through those over-estimated areas.
Note that, although the marks are drawn locally, the improvement happens globally, as the sword behind the shelf being identified and the pointed area behind the tree being refined.
Furthermore, an intentionally wrong mark outlined in yellow on the grass of the second sample does not affect the result.

The second instance shows an example of using EWDC based on extra assumptions.
Inspired by the global assumption of Berman~\emph{et~al.}~\cite{berman2016non}, we make a simplified one which assumes that pixels with similar colors should have similar transmissions if they are widely exist in the image.
In practise, pixels are firstly clustered in RGB-space.
Then, for each cluster with size large enough, a message is constructed, in which $C$ stores the pixels in the cluster and $t_s=\max_{y\in{C}}b(y)$.
An example is shown in Fig.~\ref{fig:EWDC-assume}, where EWDC successfully outlines the fence just like Berman~\emph{et~al.}~\cite{berman2016non}.

EWDC provides an interface for manual operations or other theories.
Any technology that could be transformed into the form of $C$ and $t_s$ can be used to tune the results.
The quality of EWDC depends on the quality of extra messages.
The improvements introduced by the hand-drawn messages is of high quality, while the ones introduced by the crude assumption is less accurate.

EWDC also demonstrates the advantage of WDC from an unique view.
There can be a large number of factors affecting the quality of transmission estimation,
however, WDC gathers them all into one thing, which is, the locations of dark pixels.

\section{Conclusion}
\label{sec:CON}
Dark channel prior is a highly valued theory,
however, an overlooked fact is that the dark channel of an image is actually not a fixed thing,
which changes along with the definition of local constant assumption,
and such assumption always causes defects.

In this paper, we overcome this limitation by splitting the concept of dark channel into two parts, that is, dark pixels and local constant assumption.
A novel weight map is introduced to improve the stability of the second part,
and then, WDC and CWDC are proposed, which show significant superiority in the comparisons.
Better yet, such improvement does not come along with new burdens, but simplifications.
EWDC shows that the method can be ever-improved by providing better dark pixel positions,
and the method is robust to any other changes in initial estimates.

\ifCLASSOPTIONcompsoc
\else
  \section*{Acknowledgment}
\fi

\ifCLASSOPTIONcaptionsoff
  \newpage
\fi




\bibliographystyle{IEEEtran}
\bibliography{egbib}
%
\end{document}